\documentclass{article} 
\usepackage{iclr2019_conference,times}

\usepackage{graphicx}
\usepackage{amsmath,amssymb} 
\usepackage{color}

\usepackage{times}
\usepackage{epsfig}
\usepackage{graphicx}
\usepackage{amsmath}
\usepackage{amssymb}
\usepackage{pifont} 
\usepackage{bbm}
\usepackage{empheq}
\usepackage{rotating}
\usepackage{amsthm}

\usepackage{bm}
\usepackage{siunitx}
\usepackage{textcomp}
\usepackage{subcaption}

\usepackage{caption}
\usepackage{subcaption}
\usepackage{algorithm,algpseudocode,multicol}
\usepackage{multirow}
\usepackage{makecell}
\usepackage{arydshln}
\usepackage{booktabs}
\usepackage{wrapfig}
\usepackage{listings}
\graphicspath{ {images/} }

\newcommand{\rulesep}{\unskip\ \vrule\ }

\usepackage{xspace}
\makeatletter
\DeclareRobustCommand\onedot{\futurelet\@let@token\@onedot}
\def\@onedot{\ifx\@let@token.\else.\null\fi\xspace}

\def\wrt{w.r.t\onedot} 

\makeatother

\newcommand{\real}{\mathbb{R}}
\newcommand{\dataset}{\mathcal{D}}
\newcommand{\epsmem}{\mathcal{M}}
\newcommand{\param}{\theta}

\newcommand{\ip}{\mathbf{x}}
\newcommand{\ipspace}{\mathcal{X}}
\newcommand{\descspace}{\mathcal{T}}

\newcommand{\op}{\mathbf{y}}
\newcommand{\opspace}{\mathcal{Y}}

\newcommand\numberthis{\addtocounter{equation}{1}\tag{\theequation}}

\newcommand{\SKIP}[1]{}

\newcommand{\famm}{\textsc{LCA}\xspace}
\newcommand{\mtask}{\textsc{multi-task}\xspace}
\newcommand{\mtaskz}{\textsc{multi-task-je}\xspace}
\newcommand{\van}{\textsc{van}\xspace}
\newcommand{\icarl}{\textsc{icarl}\xspace}

\newcommand{\ewc}{\textsc{ewc}\xspace}
\newcommand{\pii}{\textsc{pi}\xspace}

\newcommand{\mas}{\textsc{mas}\xspace}
\newcommand{\pnn}{\textsc{prog-nn}\xspace}
\newcommand{\rwalk}{\textsc{rwalk}\xspace}
\newcommand{\pn}{\textsc{progressive nets}\xspace}
\newcommand{\gem}{\textsc{gem}\xspace}
\newcommand{\sgem}{\textsc{a-gem}\xspace}
\newcommand{\mgem}{\textsc{s-gem}\xspace}

\newcommand{\zst}{\textsc{-je}\xspace}

\newcommand{\vanz}{\textsc{van-je}\xspace}
\newcommand{\ewcz}{\textsc{ewc-je}\xspace}
\newcommand{\piiz}{\textsc{pi-je}\xspace}
\newcommand{\masz}{\textsc{mas-je}\xspace}
\newcommand{\rwalkz}{\textsc{rwalk-je}\xspace}
\newcommand{\sgemz}{\textsc{a-gem-je}\xspace}

\newcommand\Mark[1]{\textsuperscript#1}

\usepackage[acronym,smallcaps,nowarn,section,nogroupskip,nonumberlist]{glossaries}
\glsdisablehyper{}
\newacronym{IL}{il}{incremental learning}
\newacronym{EWC}{ewc}{elastic weight consolidation}
\newacronym{PI}{pi}{path integral}
\newacronym{EWC++}{ewc++}{fast and online version of EWC}

\algnewcommand{\LeftComment}[1]{\Statex #1}

\usepackage{amsmath,amsfonts,bm}

\def\eqref#1{equation~\ref{#1}}

\def\1{\bm{1}}

\DeclareMathAlphabet{\mathsfit}{\encodingdefault}{\sfdefault}{m}{sl}
\SetMathAlphabet{\mathsfit}{bold}{\encodingdefault}{\sfdefault}{bx}{n}

\usepackage{hyperref}
\usepackage{url}

\title{Efficient Lifelong Learning with A-GEM}

\author{Arslan Chaudhry\Mark{1}, Marc'Aurelio Ranzato\Mark{2}, Marcus Rohrbach\Mark{2}, Mohamed Elhoseiny\Mark{2} \\
\Mark{1}University of Oxford, \Mark{2}Facebook AI Research\\
\texttt{arslan.chaudhry@eng.ox.ac.uk, \{ranzato,mrf,elhoseiny\}@fb.com} \\
}

\usepackage[colorinlistoftodos,prependcaption,textsize=tiny]{todonotes}
\iclrfinalcopy 
\begin{document}
\lstset{language=Python} 

\maketitle

\begin{abstract}
In lifelong learning, the learner is presented with a sequence of tasks, incrementally building a data-driven prior which may be leveraged to speed up learning of
a new task. In this work, we investigate the {\em efficiency} of current lifelong approaches, in terms of sample complexity, computational and memory cost.
Towards this end, we first introduce a new and a more realistic evaluation protocol, whereby learners observe each example only once and hyper-parameter selection is done
on a small and disjoint set of tasks, which is not used for the actual learning experience and evaluation. Second, we introduce a new metric measuring how quickly a learner 
acquires a new skill. 
Third, we propose an improved version of GEM \citep{lopez2017gradient}, dubbed Averaged GEM (\sgem), which enjoys the same or even better performance as GEM, while being
almost as computationally and memory efficient as EWC \citep{Kirkpatrick2016EWC} and other regularization-based methods. Finally, we show that all algorithms including \sgem can learn even more 
quickly if they are provided with task descriptors specifying the classification tasks under consideration. 
Our experiments on several standard lifelong learning benchmarks demonstrate that \sgem has the best trade-off between accuracy and efficiency.\footnote{The code is available at \url{https://github.com/facebookresearch/agem}.}

\end{abstract}

\section{Introduction} \label{sec:introduction}
Intelligent systems, whether they are natural or artificial, must be able to quickly adapt to changes in the environment and to quickly learn new skills
by leveraging past experiences. While current learning algorithms can achieve excellent performance on a variety of tasks, they strongly rely on copious 
amounts of supervision in the form of labeled data.

The {\em lifelong learning} (LLL) setting  attempts at addressing this shortcoming, bringing machine learning closer to a more realistic human learning by acquiring new skills quickly with a small amount of training data, given the experience accumulated in the past. In this setting, the 
learner is presented with a stream of tasks whose relatedness is not known a priori. The learner has then the potential to learn more quickly a new task,
if it can remember how to combine and re-use knowledge acquired while learning related tasks of the past. Of course, for this learning setting to be 
 useful, the model needs to be constrained in terms of amount of compute and memory required. Usually this means that the learner should not be
allowed to merely store all examples seen in the past (in which case this reduces the lifelong learning problem to a multitask problem) nor should the learner be engaged in computations
that would not be feasible in real-time, as the goal is to quickly learn from a stream of data.  

Unfortunately, the established training and evaluation protocol as well as current algorithms for lifelong learning do not satisfy all the above desiderata, 
namely learning from a stream of data using limited number of samples, limited memory and limited compute. In the most popular training paradigm, the
learner does several passes over the data \citep{Kirkpatrick2016EWC,aljundi2017memory,rusu2016progressive, progresscompress}, 
while ideally the model should need only a handful of samples and these should be provided one-by-one in a single pass~\citep{lopez2017gradient}.
 Moreover, when the learner has several hyper-parameters to tune, the current practice is to go over the sequence of tasks 
several times, each time with a different hyper-parameter value, again ignoring the requirement of learning from a stream of data and, strictly speaking, violating the assumption of the LLL scenario.
While some algorithms may work well in a single-pass setting, they unfortunately require a lot of computation \citep{lopez2017gradient} or 
their memory scales with the number of tasks \citep{rusu2016progressive},
 which greatly impedes their actual deployment in practical applications.

In this work, we propose an evaluation methodology and an algorithm that better match our desiderata, namely learning efficiently -- in terms of training samples, time and memory -- from a stream of tasks.
 First, we propose a new learning paradigm, whereby the learner
performs cross validation on a set of tasks which is disjoint from the set of tasks actually used for evaluation (Sec. \ref{sec:setup}). In this setting, the learner will have to learn and will
be tested on an entirely new sequence of tasks and it will perform just a single pass over this data stream. Second, we build upon GEM~\citep{lopez2017gradient}, an algorithm which leverages a small
episodic memory to perform well in a single pass setting, and propose a small change to the loss function which makes GEM orders of magnitude faster at training time while maintaining similar performance; we dub this variant of GEM, \sgem (Sec. \ref{sec:approach}). Third, we explore the use of compositional task descriptors in order 
to improve the few-shot learning performance within LLL showing that with this additional information the learner can pick up new skills
more quickly (Sec. \ref{sec:joint_embed}). 
Fourth, we introduce a new metric to measure the speed of learning, which is useful to quantify the ability of a learning algorithm to learn a new task (Sec.~\ref{sec:metrics}).
And finally, using our new learning paradigm and metric, we demonstrate \sgem on a variety of benchmarks and against several representative baselines (Sec. \ref{sec:experiments}). 
Our experiments show that \sgem has a better trade-off between average accuracy and computational/memory cost. 
Moreover, all algorithms improve their ability to quickly learn a new task when provided with compositional task descriptors, and they do so better and better as they progress through the learning experience.

\section{Learning Protocol} \label{sec:setup}
\vspace{-2mm}
Currently, most works on lifelong learning~\citep{Kirkpatrick2016EWC, rusu2016progressive, shin2017continual, nguyen2017variational} adopt a learning protocol which is directly borrowed from supervised learning. There are $T$ tasks, and each task
consists of a training, validation and test sets. During training the learner does as many passes over the data of each task as desired. Moreover, hyper-parameters are tuned
on the validation sets by sweeping over the whole sequence of tasks as many times as required by the cross-validation grid search. Finally, metrics of interest are reported
on the test set of each task using the model selected by the previous cross-validation procedure.

Since the current protocol violates our stricter definition of LLL for which the learner can only make a single pass over the data, as we want to 
emphasize the importance of learning quickly from data, we now introduce a new learning protocol. 

We consider two streams of tasks, described by the following ordered sequences of datasets $\dataset^{CV} = \{\dataset_1, \cdots,\dataset_{T^{CV}}\}$ and 
$\dataset^{EV} = \{\dataset_{{T^{CV}}+1}, \cdots,\dataset_T\}$, 
where $\dataset_k = \{(\ip_i^k, t_i^k, y_i^k)_{i=1}^{n_k}\}$ is the dataset of the $k$-th task, ${T^{CV}} < T$ (in all our experiments $T^{CV}=3$ while $T=20$), 
and we assume that all datasets are drawn from the same distribution over tasks. To avoid cluttering of the notation, we let the context specify whether 
$\dataset_k$ refers to the training or test set of the $k$-th dataset. 

$\dataset^{CV}$ is the stream of datasets which will be used during cross-validation;
$\dataset^{CV}$ allows the learner to replay all samples multiple times for the purposes of model hyper-parameter selection. 
Instead, $\dataset^{EV}$ is the actual
dataset used for final training and evaluation on the test set; the learner will observe training examples from $\dataset^{EV}$ once and only once, and all metrics will be 
reported on the test sets of $\dataset^{EV}$. Since the regularization-based approaches for lifelong learning~\citep{Kirkpatrick2016EWC,Zenke2017Continual} are rather sensitive to the choice of the regularization hyper-parameter, 
we introduced the set $\dataset^{CV}$, as it seems reasonable in practical applications to have similar tasks that can be used for tuning the system. However, the actual
training and testing are then performed on $\dataset^{EV}$ using a single pass over the data.
See Algorithm~\ref{algo:protocol} for a summary of the training and evaluation protocol.

Each example in any of these dataset consists of a triplet defined by an input ($\ip^k \in \ipspace$), 
task descriptor ($t^k \in \descspace$, see Sec. \ref{sec:joint_embed} for examples) and a target vector ($y^k \in \op^k$), 
where $\op^k$ is the set of labels specific to task $k$ and $\op^k \subset \opspace$. 
 While observing the data, the goal is to learn a predictor $f_{\param}: \ipspace \times \descspace \rightarrow \opspace$, 
parameterized by $\param \in \real^P$ (a neural network  in our case), that can map any test pair $(\ip, t)$ to a target $y$.

\begin{algorithm}[t]
\footnotesize
\caption{Learning and Evaluation Protocols\label{algo:protocol}}
\begin{algorithmic}[1]
  \For{$h$ \textbf{in} hyper-parameter list} \Comment{Cross-validation loop, executing multiple passes over $\dataset^{CV}$}
    \For{$k=1$ \textbf{to} $T^{CV}$} \Comment{Learn over data stream $\dataset^{CV}$ using $h$}
      \For{$i=1$ \textbf{to} $n_k$} \Comment{Single pass over  $\dataset_k$}
        \State Update $f_\theta$ using  $(\ip_i^k, t_i^k, y_i^k)$ and hyper-parameter $h$
        \State Update metrics on test set of $\dataset^{CV}$
        \EndFor
        \EndFor
        \EndFor
  \State Select best hyper-parameter setting, $h^*$, based on average accuracy of test set of $\dataset^{CV}$, see Eq.~\ref{eq:avg_acc}.
  \State Reset $f_\theta$.
  \State Reset all metrics.
  \For{$k=T^{CV}+1$ \textbf{to} $T$} \Comment{Actual learning over datastream $\dataset^{EV}$}
    \For{$i=1$ \textbf{to} $n_k$} \Comment{Single pass over  $\dataset_k$}
      \State Update $f_\theta$ using  $(\ip_i^k, t_i^k, y_i^k)$ and hyper-parameter $h^*$
      \State Update metrics on test set of $\dataset^{EV}$
  \EndFor
  \EndFor
  \State Report metrics on test set of $\dataset^{EV}$.
\end{algorithmic}
\end{algorithm}

\section{Metrics} \label{sec:metrics}
Below we describe the metrics used to evaluate the LLL methods studied in this work. 
In addition to Average Accuracy ($A$) and Forgetting Measure ($F$)~\citep{chaudhry2018riemannian}, 
we define a new measure, the Learning Curve Area (\famm), that captures how quickly a model learns.

The training dataset of each task, $\dataset_k$, consists of a total $B_k$ mini-batches. 
After each presentation of a mini-batch of task $k$, we evaluate the performance of the learner on all the tasks using the corresponding test sets.
 Let $a_{k,i,j} \in [0,1]$ be the accuracy evaluated on the test set of task
$j$, after the model has been trained with the $i$-th mini-batch of task $k$. Assuming the first learning task in the continuum is indexed by $1$ (it will be $T^{CV}+1$ for $\dataset^{EV}$) and the last one by $T$ (it will be $T^{CV}$ 
for $\dataset^{CV}$),  we define the following metrics:

\paragraph{Average Accuracy ($A \in [0,1]$)} Average accuracy after the model has been trained continually with all the mini-batches up till task $k$ is defined as:
\begin{equation}
        \label{eq:avg_acc}
        A_k = \frac{1}{k} \sum_{j=1}^k a_{k, B_k, j}
\end{equation}
In particular, $A_T$ is the average accuracy on all the tasks after the last task has been learned; this is the most commonly used metric used in LLL.

\paragraph{Forgetting Measure ($F \in [-1,1]$)}\citep{chaudhry2018riemannian} Average forgetting after the model has been trained continually with all the mini-batches up till task $k$ is defined as:
\begin{equation}
        \label{eq:fgt}
        F_k = \frac{1}{k-1} \sum_{j=1}^{k-1} f_j^k
\end{equation}
where $f_j^k$ is the forgetting on task `$j$' after the model is trained with all the mini-batches up till task $k$ and computed as:
\begin{equation}
        f_j^k = \max_{l \in \{1,\cdots,k-1\}} a_{l, B_{l}, j} - a_{k, B_k, j}
\end{equation}
Measuring forgetting after all tasks have been learned is important for a two-fold reason. It quantifies the accuracy drop on past tasks, and it 
gives an indirect notion of how quickly a model may learn a new task, since a forgetful model will have little knowledge left to transfer, particularly so if the new
task relates more closely to one of the very first tasks encountered during the learning experience.

\paragraph{Learning Curve Area ($\famm \in [0,1]$)} Let us first define an average $b$-shot performance (where $b$ is the mini-batch number) after the model has been trained for all the $T$ tasks as:
\begin{equation}
        \label{eq:k_shot_perf}
        Z_b = \frac{1}{T} \sum_{k=1}^T a_{k,b,k}
\end{equation}

\famm\ at $\beta$ is the area of the convergence curve $Z_b$ as a function of $b \in [0, \beta]$:
\begin{equation}
        \label{eq:fam}
        \famm_{\beta} = \frac{1}{\beta+1}\int_{0}^{\beta} Z_b db = \frac{1}{\beta+1}\sum_{b=0}^\beta Z_b
\end{equation}
\famm has an intuitive interpretation. $\famm_{0}$ is the average 0-shot performance, the same as forward transfer in~\citet{lopez2017gradient}. $\famm_{\beta}$ is the area under the $Z_b$ curve, which
is high if the 0-shot performance is good and if the learner learns quickly. In particular, there could be two models with the same $Z_{\beta}$ or $A_T$, but very different
$\famm_{\beta}$ because one learns much faster than the other while they both eventually obtain the same final accuracy. This metric aims at discriminating between these two cases, and it makes sense for relatively small values of $\beta$ since we are interested in models that learn from few examples.

\section{Averaged Gradient Episodic Memory (\sgem)}
\label{sec:approach}
So far we discussed a better training and evaluation protocol for LLL and a new metric to measure the speed of learning. Next, we review \gem~\citep{lopez2017gradient}, which is an algorithm that has been shown to work well in the single epoch setting. Unfortunately, \gem is very intensive in terms of computational and memory cost,
which motivates our efficient variant, dubbed \sgem. In Sec.~\ref{sec:joint_embed}, we will describe how 
compositional task descriptors can be leveraged to further speed up learning in the few shot regime. 

\gem\ avoids catastrophic forgetting by storing an episodic memory $\epsmem_k$ for each task $k$. While minimizing the loss on the current task $t$, \gem\ treats the losses on the episodic memories of tasks $k < t$, given by $\ell(f_{\param}, \epsmem_k)~=~\frac{1}{|\epsmem_k|} \sum_{(\ip_i, k, y_i) \in \epsmem_k} \ell(f_{\param}(\ip_i, k), y_i)$, as inequality constraints, avoiding their increase but allowing their decrease. This effectively permits \gem\ to do positive backward transfer which other LLL methods do not support. Formally, at task $t$, \gem\ solves for the following objective: 
\begin{equation}
	\textrm{minimize}_{\param}  \quad  \ell(f_{\param}, \dataset_t) \quad \textrm{s.t.} \quad  \ell(f_{\param}, \epsmem_k) \leq \ell(f_{\param}^{t-1}, \epsmem_k) \quad\quad \forall k < t \quad \numberthis \label{eq:gem_main_obj}
\end{equation}	 
Where $f_{\param}^{t-1}$ is the network trained till task $t-1$. To inspect the increase in loss, \gem\ computes the angle between the loss gradient vectors of previous tasks $g_k$, and the proposed gradient update on the current task $g$. Whenever the angle is greater than $\ang{90}$ with any of the $g_k$'s, it projects the proposed gradient to the closest in L2 norm gradient $\tilde{g}$ that keeps the angle within the bounds. Formally, the optimization problem \gem\ solves is given by:
\begin{equation}
	 \textrm{minimize}_{\tilde{g}} \quad \frac{1}{2}||g-\tilde{g}||_2^2 \quad \textrm{s.t.} 	\quad \langle\tilde{g}, g_k\rangle \geq 0 \quad \quad\quad\quad\quad\quad\quad\forall k < t \numberthis \label{eq:gem_primal_obj}
\end{equation}
Eq.\ref{eq:gem_primal_obj} is a quadratic program (QP) in $P$-variables (the number of parameters in the network), which for neural networks could be in millions. 
In order to solve this efficiently, \gem\ works in the dual space which results in a much smaller QP with only $t-1$ variables:
\begin{equation}
	 \textrm{minimize}_v \quad \frac{1}{2} v^\top GG^\top v + g^\top G^\top v \quad \textrm{s.t.} 				\quad v\geq 0\quad\quad\quad\quad\quad\quad \quad\quad\numberthis \label{eq:gem_dual_obj}
\end{equation}
where $G=-(g_1,\cdots,g_{t-1}) \in \real^{(t-1) \times P}$ is computed at each gradient step of training. Once the solution $v^*$ to Eq.~\ref{eq:gem_dual_obj} is found, the projected gradient update can be computed as $\tilde{g}=G^\top v^*+g$.

While \gem\ has proven very effective in a single epoch setting~\citep{lopez2017gradient}, the performance gains come at a big computational burden at training time.
 At each training step, \gem\ computes the matrix $G$ using all samples from the episodic memory, and it also needs to solve the QP of Eq.~\ref{eq:gem_dual_obj}. 
Unfortunately, this inner loop optimization becomes prohibitive when the size of $\epsmem$ and the number of tasks is large, see Tab.~\ref{tab:gem_vs_sgem_time} in Appendix for an empirical
analysis.  To alleviate the computational burden of \gem, next we propose a much more efficient version of \gem, called Averaged GEM (\sgem).

Whereas \gem\ ensures that at every training step the loss of each {\em individual} previous tasks, approximated by the samples in episodic memory, does not increase, \sgem\ 
tries to ensure that at every training step the {\em average} episodic memory loss over the previous tasks does not increase. 
Formally, while learning task $t$, the objective of \sgem\ is:
\begin{align*}
	\textrm{minimize}_{\param}  \quad & \ell(f_{\param}, \dataset_t) \quad \textrm{s.t.} 	\quad \ell(f_{\param}, \epsmem) \leq \ell(f_{\param}^{t-1}, \epsmem) \quad\quad \textrm{where }  \epsmem = \cup_{k<t} \epsmem_k \numberthis \label{eq:s_gem_main_obj}	
\end{align*}
The corresponding optimization problem reduces to:
\begin{align*}
	 	\textrm{minimize}_{\tilde{g}} \quad &\frac{1}{2}||g-\tilde{g}||_2^2  \quad \textrm{s.t.} 				\quad \tilde{g}^\top g_{ref} \geq 0  \numberthis \label{eq:s_gem_primal_obj}
\end{align*}
where $g_{ref}$ is a gradient computed using a batch randomly sampled from the episodic memory, $(\ip_{ref}, y_{ref}) \sim \epsmem$, of all the past tasks. 
In other words, \sgem\ replaces the $t-1$ constraints of \gem\ with a single constraint, where $g_{ref}$ is the average of the gradients from the previous tasks
computed from a random subset of the episodic memory.

The constrained optimization problem of Eq.~\ref{eq:s_gem_primal_obj} can now be solved very quickly; when the gradient $g$ violates the constraint, it is projected via:
\begin{equation}
\label{eq:agem_update_rule}
\tilde{g} = g - \frac{g^\top g_{ref}}{g_{ref}^\top g_{ref}}g_{ref}  \end{equation}
The formal proof of the update rule of \sgem\ (Eq.~\ref{eq:agem_update_rule}) is given in Appendix~\ref{sec:supp_sgem_update_rule}. This makes \sgem\ not only memory efficient, as it does not need to store the  matrix $G$, but also orders of magnitude faster than \gem\ because 1) 
it is not required to compute the matrix $G$ but just the gradient of a random subset of memory examples, 2) it does not need to solve any QP but just an inner product, and 3)
it will incur in less violations particularly when the number of tasks is large (see Tab.~\ref{tab:gem_vs_sgem_time} and Fig.~\ref{fig:gem_vs_sgem_constrnt_violtn} in Appendix for empirical evidence).
All together these factors make \sgem\ faster while not hampering its good performance in the single pass setting. 

Intuitively, the difference between \gem\ and \sgem\ loss
 functions is that \gem\ has better guarantess in terms of worst-case forgetting of each individual task since (at least on the memory examples) it prohibits an increase of any task-specific loss, while \sgem\ has better guaratees in terms of average accuracy since \gem\ may prevent a gradient step because of a task constraint violation 
although the overall average loss may actually decrease, see Appendix Sec.~\ref{sec:supp_freq_const_violtn} and \ref{sec:avg_Acc_wst_fgt} for further analysis and empirical evidence.
 The pseudo-code of \sgem\ is given in Appendix Alg.~\ref{alg:sgem}. 

\section{Joint Embedding Model Using Compositional Task Descriptors} \label{sec:joint_embed}
In this section, we discuss how we can improve forward transfer for all the LLL methods including \sgem. In order to speed up learning of a new task, we consider the use of compositional task descriptors where components are shared across tasks and thus allow transfer. Examples of compositional task descriptors are, for instance, 
 a natural language description of the task under consideration or a matrix specifying the attribute values of the objects to be recognized in the task.
In our experiments, we use the latter since it is provided with popular benchmark datasets~\citep{WahCUB_200_2011,lampert2009learning}.
For instance, if the model has already learned and remembers about two independent properties (e.g., color of feathers and shape of beak), it can quickly recognize
a new class provided a descriptor specifying the values of its attributes (yellow feathers and red beak), although this is an entirely unseen combination.

Borrowing ideas from literature in few-shot learning~\citep{lampert2014attribute,zhang2018large,elhoseiny2017write,xian2018zero}, we learn a joint embedding space between image features and the attribute embeddings.
Formally, let $\ip^k \in \ipspace$ be the input (e.g., an image), $t^k$ be the task descriptor in the form of a matrix of size $C_k \times A$, where $C_k$ is the number of classes
in the $k$-th task and $A$ is the total number of attributes for each class in the dataset. The joint embedding model consists of
a feature extraction module, $\phi_{\param}: \ip^k \rightarrow \phi_{\param}(\ip^k)$, where $\phi_{\param}(\ip^k) \in \real^D$, 
and a task embedding module, $\psi_{\omega}: t^k \rightarrow \psi_{\omega}(t^k)$, where $\psi_{\omega}(t^k) \in \real^{C_k \times D}$. 
In this work, $\phi_{\param}(.)$ is implemented as a standard multi-layer feed-forward network (see Sec.~\ref{sec:experiments} for the exact parameterization), whereas $\psi_{\omega}(.)$ is implemented as a parameter matrix of dimensions $A \times D$. 
This matrix can be interpreted as an attribute look-up table as each attribute is associated with a $D$ dimensional vector, from which a class embedding vector is constructed via a linear combination of the attributes present in the class; the task descriptor embedding is then the concatenation of the embedding vectors of the classes present in the task (see Appendix Fig.~\ref{fig:je_pict_desc} for the pictorial description of the joint embedding model). During training, the parameters $\param$ and $\omega$ are learned by minimizing the cross-entropy loss:
\begin{equation}
        \ell_k(\param, \omega) =  \frac{1}{N} \sum_{i=1}^N-\log(p(y_i^k | \ip_i^k, t^k; \theta, \omega))
\end{equation}
where ($\ip_i^k, t^k, y_i^k$) is the $i$-th example of task $k$. If $y_i^k=c$, then the distribution $p(.)$ is given by:
\begin{equation}
        p(c | \ip_i^k, t^k; \theta, \omega) = \frac{\exp([\phi_{\theta}(\ip_i^k)\psi_{\omega}(t^k)^\top]_c)}{\sum_j \exp([\phi_{\theta}(\ip_i^k)\psi_{\omega}(t^k)^\top]_j)}
\end{equation}
where $[a]_i$ denotes the $i$-th element of the vector $a$. Note that the architecture and loss functions are general, 
and apply not only to \sgem but also to any other LLL model (e.g., regularization based approaches). See Sec.~\ref{sec:experiments} for the actual choice of 
parameterization of these functions.

\section{Experiments} \label{sec:experiments}

We consider four dataset streams, see Tab.\ref{tab:datasets} in Appendix Sec.~\ref{sec:dataset_stats} for a summary of the statistics. 
\textbf{Permuted MNIST}~\citep{Kirkpatrick2016EWC} is  a variant of MNIST~\citep{lecun1998mnist} dataset of handwritten digits where 
each task has a certain random permutation of the input pixels which is applied to all the images of that task.
\textbf{Split CIFAR}~\citep{Zenke2017Continual} consists of splitting the original CIFAR-100 dataset~\citep{krizhevsky2009learning} into $20$ disjoint 
subsets, where each subset is constructed by randomly sampling $5$ classes {\em without} replacement from a total of $100$ classes. Similarly to Split CIFAR, \textbf{Split CUB} is an incremental version of the fine-grained image classification dataset CUB~\citep{WahCUB_200_2011} of $200$ bird categories split into $20$ disjoint subsets of classes. \textbf{Split AWA}, on the other hand, is the incremental version of the AWA dataset~\citep{lampert2009learning} of $50$ animal categories, where each task is constructed by sampling $5$ classes {\em with} replacement from the total $50$ classes, constructing $20$ tasks. In this setting, classes may overlap among multiple tasks, but within each task they compete against different set of classes. Note that to make sure each training example is only seen once, the training data of a each class is split into disjoint sets depending on the frequency of its occurrence in different tasks. For Split AWA, the classifier weights of each class are randomly initialized within each head without any transfer from the previous occurrence of the class in past tasks. Finally, while on Permuted MNIST and Split CIFAR we provide integer task descriptors, on Split CUB and Split AWA we stack together the attributes of the classes (specifying for instance the type of beak, the color of feathers, etc.) belonging to the current task to form a descriptor.


In terms of architectures, we use a fully-connected network with two hidden layers of $256$ ReLU units each for Permuted MNIST, a reduced ResNet18 for Split CIFAR like 
in~\citet{lopez2017gradient}, and a standard  ResNet18~\citep{he2016deep} for Split CUB and Split AWA. For a given dataset stream, all models use the same architecture,
and all models are optimized via stochastic gradient descent with mini-batch size equal to 10. We refer to the joint-embedding model version of these models by appending
the suffix `\zst' to the method name. 

As described in Sec.~\ref{sec:setup} and outlined in Alg.~\ref{algo:protocol}, in order to cross validate we use the first 3 tasks, and then report metrics on
the remaining 17 tasks after doing a single training pass over each task in sequence.

Lastly, we compared \sgem against several baselines and state-of-the-art LLL approaches which we describe next. 
\textbf{\van} is a single supervised learning model, trained continually without any regularization, with the parameters of a new task initialized from the parameters of the previous task. 
\textbf{\icarl}~\citep{Rebuffi16icarl} is a class-incremental learner that uses nearest-exemplar-based classifier and avoids catastrophic forgetting by regularizing 
over the feature representation of previous tasks using a knowledge distillation loss.
\textbf{\ewc}~\citep{Kirkpatrick2016EWC}, \textbf{\pii}~\citep{Zenke2017Continual}, \textbf{\rwalk}~\citep{chaudhry2018riemannian} and
 \textbf{\mas}~\citep{aljundi2017memory} are regularization-based approaches aiming at avoiding catastrophic forgetting by limiting learning of parameters critical to
the performance of past tasks. 
\textbf{Progressive Networks} (\pnn)~\citep{rusu2016progressive} is a modular approach whereby a new ``column'' with lateral connections to previous hidden layers is added once a new task arrives.
\textbf{\gem}~\citep{lopez2017gradient} described in Sec.~\ref{sec:approach} is another natural baseline of comparison since \sgem builds upon it.
The amount of episodic memory per task used in \icarl, \gem\ and \sgem\ is set to $250$, $65$, $50$, and $100$, and the batch size for the computation of $g_{ref}$ (when the episodic memory is sufficiently filled) in \sgem\ is set to $256$, $1300$, $128$ and $128$ for MNIST, CIFAR, CUB and AWA, respectively. While populating episodic memory, the samples are chosen uniformly at random for each task. Whereas the network weights are randomly initialized for MNIST, CIFAR and AWA, on the other hand, for CUB, due to the small dataset size, a pre-trained ImageNet model is used. 
Finally, we consider a multi-task baseline, \textbf{\mtask}, trained on a single pass over shuffled data from all tasks, and thus violating the LLL assumption. It can be seen as an upper bound performance for average accuracy.
	
\subsection{Results}
\label{subsec:results}
\begin{figure}[t]
		\def \SUBWIDTH {0.11\linewidth}
		\def \FIGSCALE {0.4}
		\def \HEIGHT {10ex}
		\def \HORZSPACE {-0.25em}
        \begin{subfigure}{\SUBWIDTH}
        \begin{center}
                \includegraphics[scale=\FIGSCALE]{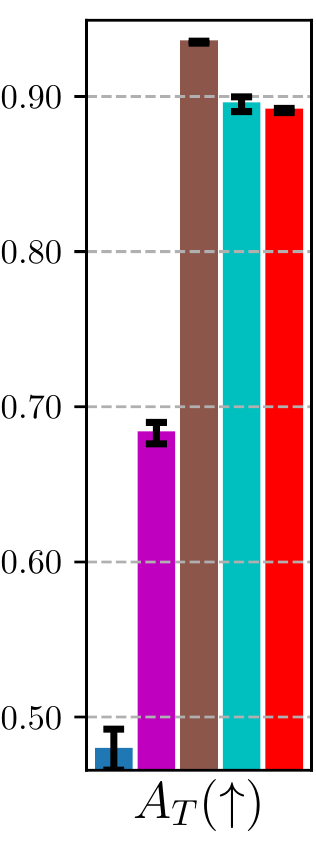}
        \end{center}
        \end{subfigure}\hspace{\HORZSPACE}
        \begin{subfigure}{\SUBWIDTH}
        \begin{center}
                \includegraphics[scale=\FIGSCALE]{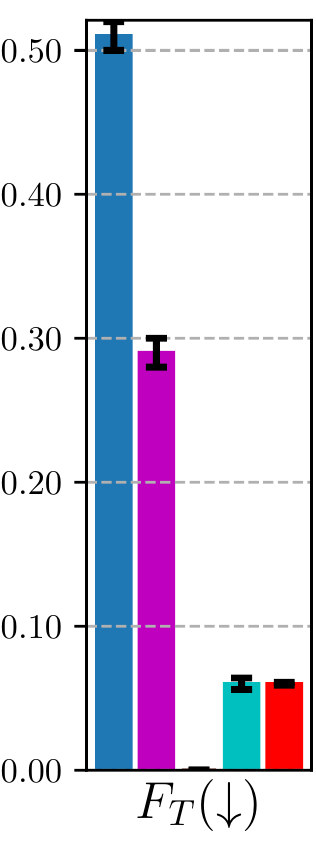}
        \end{center}
        \end{subfigure}\hspace{\HORZSPACE}
        \begin{subfigure}{\SUBWIDTH}
        \begin{center}
                \includegraphics[scale=\FIGSCALE]{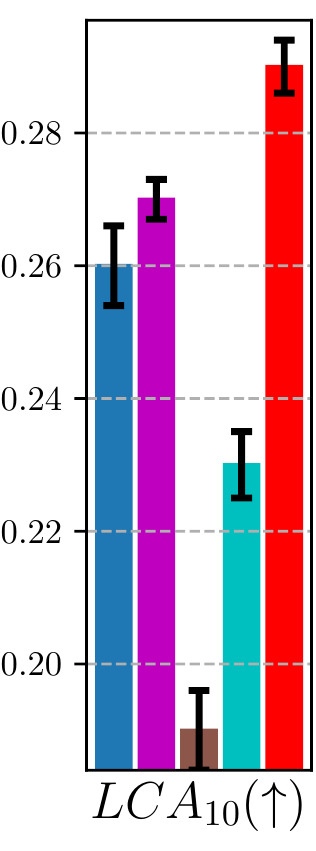}
        \end{center}
        \end{subfigure}\hspace{\HORZSPACE}
        \begin{subfigure}{\SUBWIDTH}
        \begin{center}
                \includegraphics[scale=\FIGSCALE]{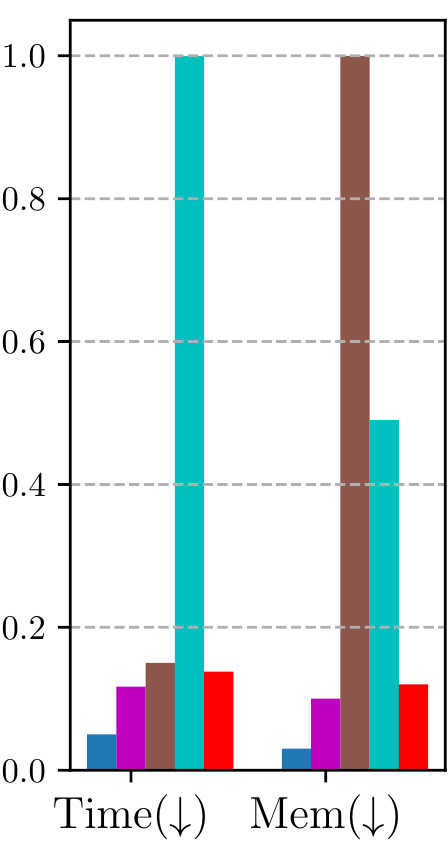}
        \end{center}
        \end{subfigure}\hspace{0.9em}
        \rulesep
        \begin{subfigure}{\SUBWIDTH}
        \begin{center}
                \includegraphics[scale=\FIGSCALE]{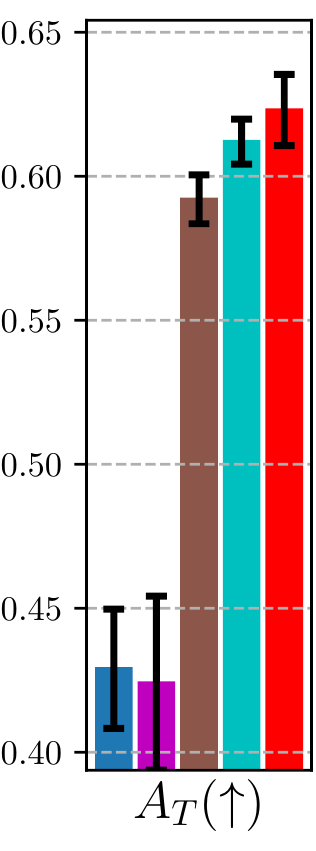}
        \end{center}
        \end{subfigure}\hspace{\HORZSPACE}
        \begin{subfigure}{\SUBWIDTH}
        \begin{center}
                \includegraphics[scale=\FIGSCALE]{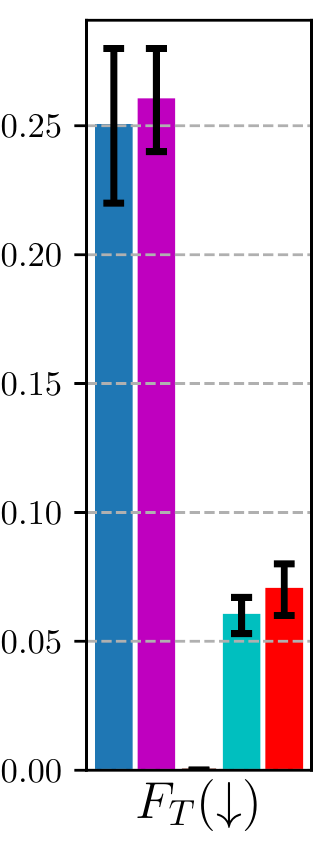}
        \end{center}
        \end{subfigure}\hspace{\HORZSPACE}
        \begin{subfigure}{\SUBWIDTH}
        \begin{center}
                \includegraphics[scale=\FIGSCALE]{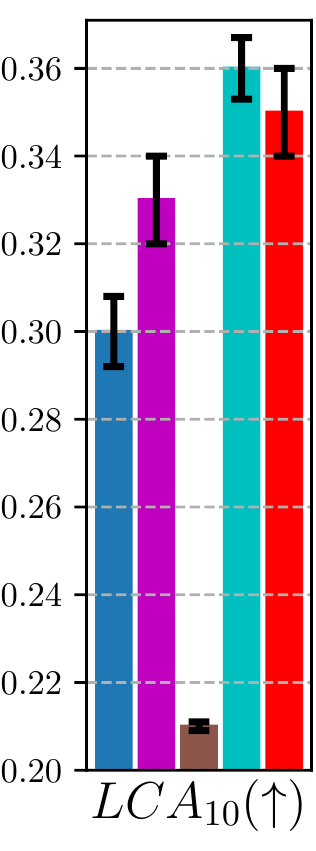}
        \end{center}
        \end{subfigure}\hspace{\HORZSPACE}
        \begin{subfigure}{\SUBWIDTH}
        \begin{center}
                \includegraphics[scale=\FIGSCALE]{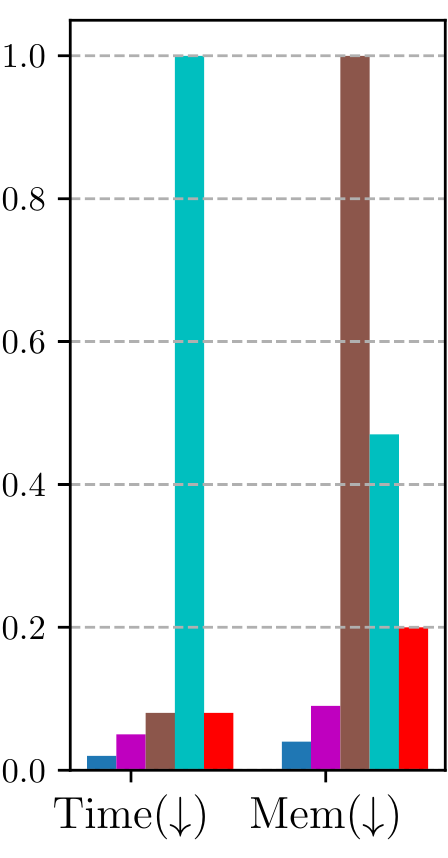}
        \end{center}
        \end{subfigure}\hspace{0.5em}
        \begin{subfigure}{0.05\linewidth}
        \begin{center}
                \includegraphics[scale=0.2]{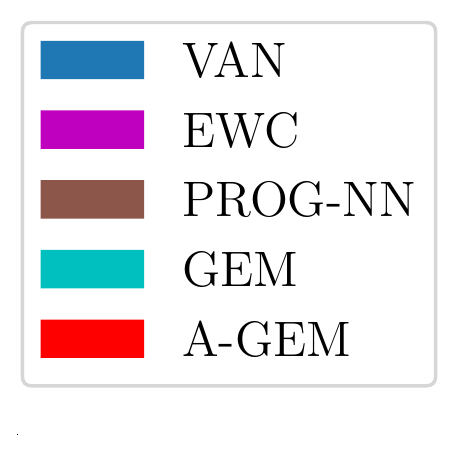}
        \end{center}
        \end{subfigure}
       
	\begin{minipage}[t]{.5\linewidth}
		\centering
		\subcaption{Permuted MNIST}
	\end{minipage}%
	\begin{minipage}[t]{.5\linewidth}
		\centering
		\subcaption{Split CIFAR}
	\end{minipage}

    \caption{\em Performance of LLL models across different measures on Permuted MNIST and Split CIFAR. 
For Accuracy ($A_T$) and Learning Curve Measure ($LCA_{10}$) the higher the number (indicated by $\uparrow$) the better is the model. For Forgetting ($F_T$), Time and Memory the lower the number (indicated by $\downarrow$) the better is the model. 
For Time and Memory, the method with the highest complexity is taken as a reference (value of $1$) and the other methods are reported relative to that method. $A_T$, $F_T$ and $LCA_{10}$ values and confidence intervals are computed over 5 runs. \sgem provides the best trade-off across different measures and dimensions. Other baselines are given in Tab.~\ref{tab:mnist_cifar_comp} and \ref{tab:gem_vs_sgem_time} in the Appendix, which are used to generate the plots.}
\label{fig:grouped_cifar_mnist}
\end{figure}

\begin{figure}[t]
         \def \SUBWIDTH {0.12\linewidth}
		\def \FIGSCALE {0.45}
		\def \HEIGHT {10ex}
		\def \HORZSPACE {-0.15em}
        \begin{subfigure}{\SUBWIDTH}
        \begin{center}
                \includegraphics[scale=\FIGSCALE]{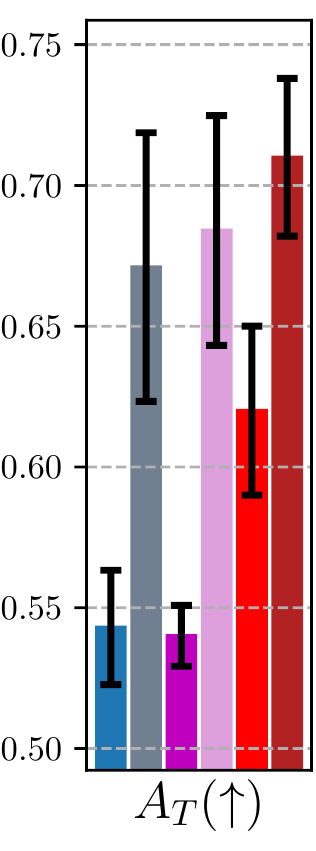}
        \end{center}
        \end{subfigure}\hspace{\HORZSPACE}
        \begin{subfigure}{\SUBWIDTH}
        \begin{center}
                \includegraphics[scale=\FIGSCALE]{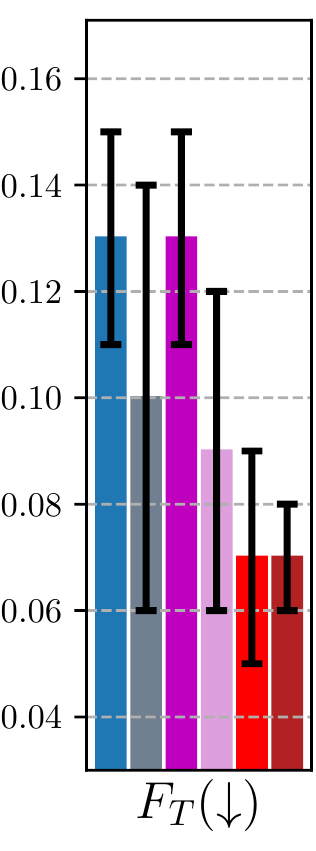}
        \end{center}
        \end{subfigure}\hspace{\HORZSPACE}
        \begin{subfigure}{\SUBWIDTH}
        \begin{center}
                \includegraphics[scale=\FIGSCALE]{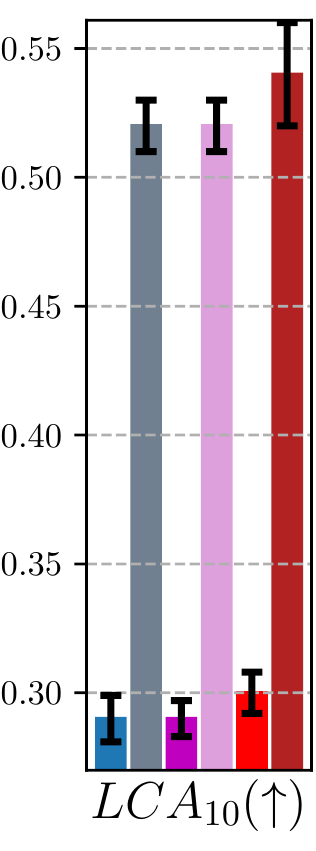}
        \end{center}
        \end{subfigure}\hspace{\HORZSPACE}
        \begin{subfigure}{0\SUBWIDTH}
        \begin{center}
                \includegraphics[scale=\FIGSCALE]{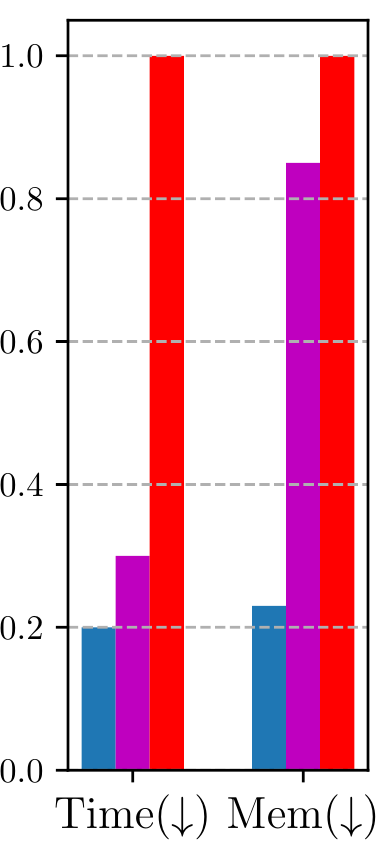}
        \end{center}
        \end{subfigure}\hspace{0.15em}
        \rulesep
        \begin{subfigure}{\SUBWIDTH}
        \begin{center}
                \includegraphics[scale=\FIGSCALE]{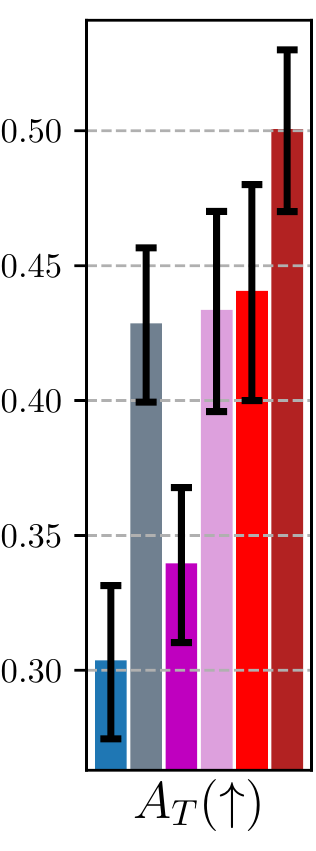}
        \end{center}
        \end{subfigure}\hspace{\HORZSPACE}
        \begin{subfigure}{\SUBWIDTH}
        \begin{center}
                \includegraphics[scale=\FIGSCALE]{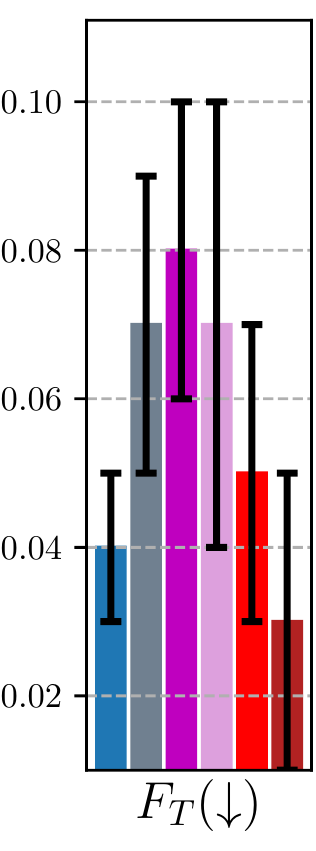}
        \end{center}
        \end{subfigure}\hspace{\HORZSPACE}
        \begin{subfigure}{\SUBWIDTH}
        \begin{center}
                \includegraphics[scale=\FIGSCALE]{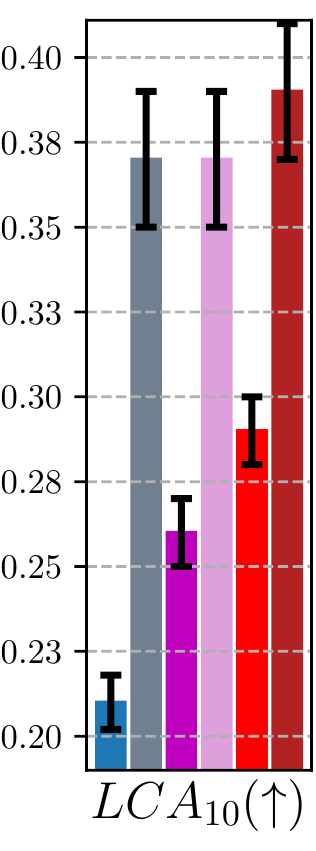}
        \end{center}
        \end{subfigure}\hspace{\HORZSPACE}
        \begin{subfigure}{\SUBWIDTH}
        \begin{center}
                \includegraphics[scale=\FIGSCALE]{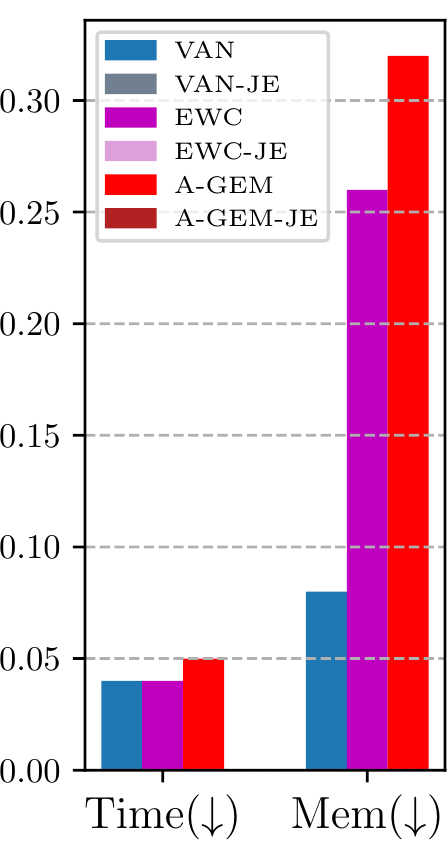}
        \end{center}
        \end{subfigure}
       
	\begin{minipage}[t]{.5\linewidth}
		\centering
		\subcaption{Split CUB}
	\end{minipage}%
	\begin{minipage}[t]{.5\linewidth}
		\centering
		\subcaption{Split AWA}
	\end{minipage}
    \caption{\em Performance of LLL models across different measures on Split CUB and Split AWA. On both the datasets, \pnn runs out of memory. The memory and time complexities of joint embedding models are the same as those of the corresponding standard models and are hence omitted. $A_T$, $F_T$ and $LCA_{10}$ values and confidence intervals are computed over 10 runs. Other baselines are given in Tab.~\ref{tab:cub_comp},~\ref{tab:awa_comp} and \ref{tab:gem_vs_sgem_time} in the Appendix, which are used to generate the plots.}
\label{fig:grouped_cub_awa}
\end{figure}

\begin{figure}[t]
	\begin{subfigure}{0.45\linewidth}
	\begin{center}
		\includegraphics[scale=0.41]{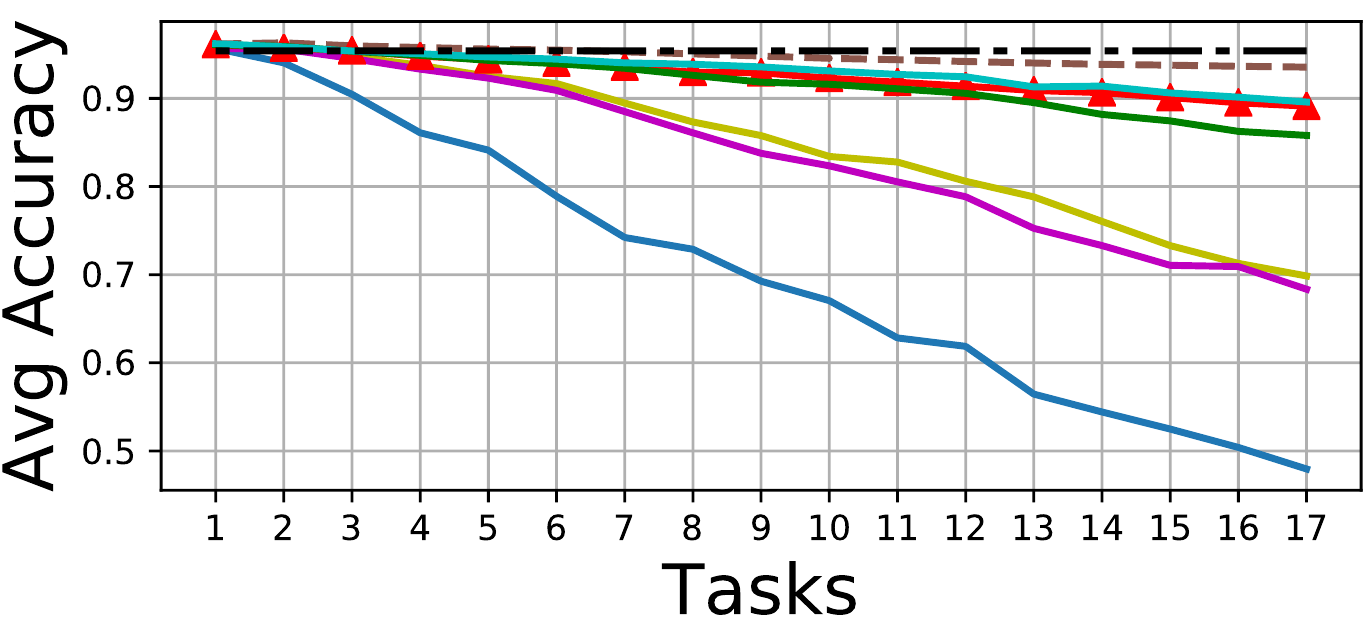}
	\end{center}
	\end{subfigure}%
	\begin{subfigure}{0.55\linewidth}
	\begin{center}
		\includegraphics[scale=0.41]{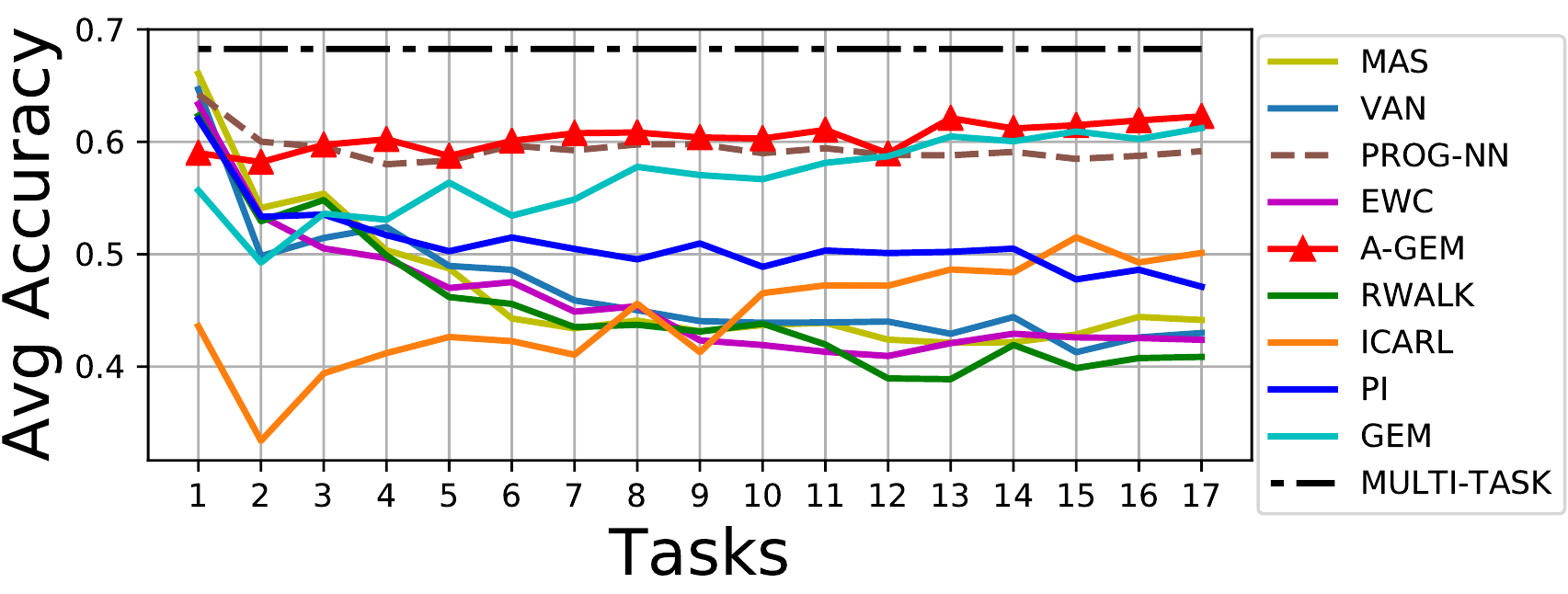}
	\end{center}
	\end{subfigure}
	\begin{subfigure}{0.45\linewidth}
	\begin{center}
		\includegraphics[scale=0.41]{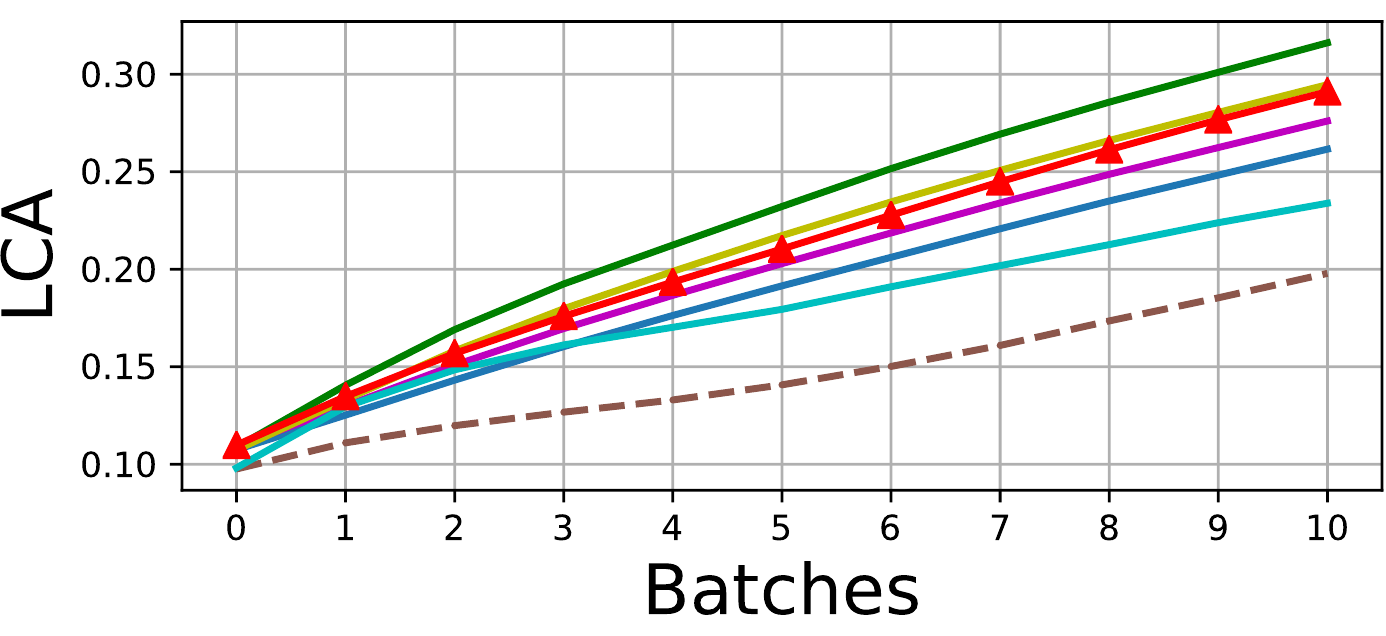}
		\caption{\small Permuted MNIST}
	\end{center}
	\end{subfigure}%
	\begin{subfigure}{0.45\linewidth}
	\begin{center}
		\includegraphics[scale=0.41]{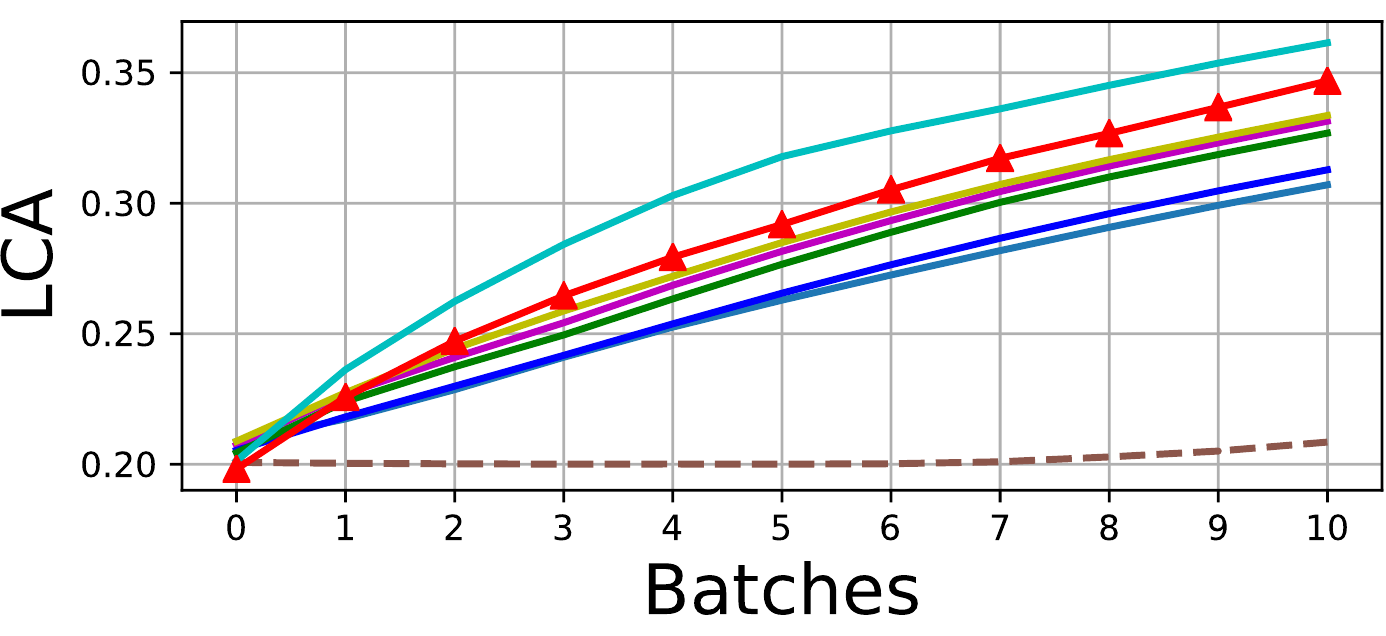}
		\caption{\small Split CIFAR}
	\end{center}
	\end{subfigure}%
\caption{\em \textbf{Top Row:} Evolution of average accuracy ($A_k$) as new tasks are learned. 
\textbf{Bottom Row:} Evolution of \famm during the first ten mini-batches. }
\label{fig:mnist_cifar_avg_acc_evol}
\end{figure}

\begin{figure}[t]
        \begin{subfigure}{0.45\linewidth}
        \begin{center}
                \includegraphics[scale=0.41]{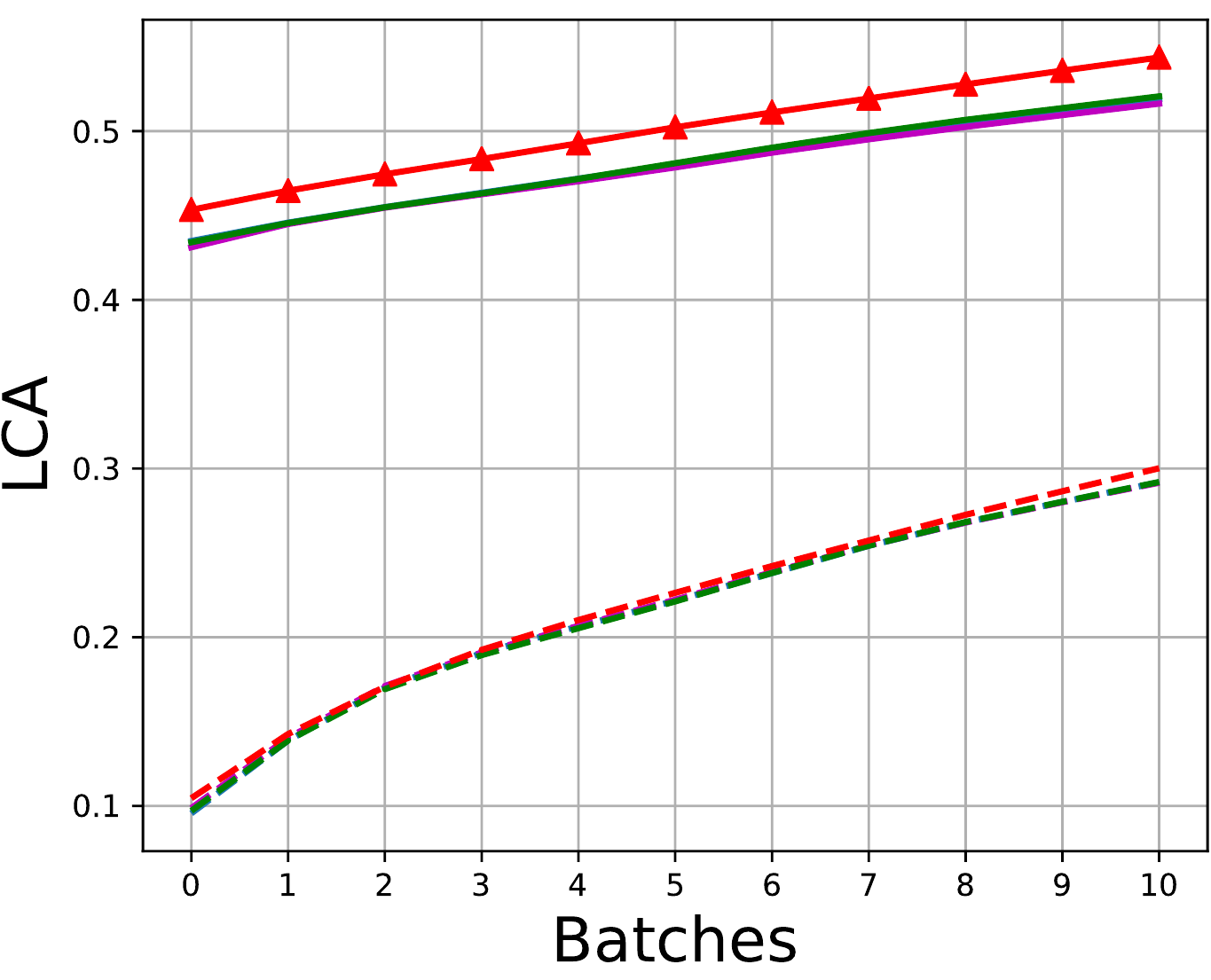}
                \caption{\small Split CUB}
        \end{center}
        \end{subfigure}%
        \begin{subfigure}{0.55\linewidth}
        \begin{center}
                \includegraphics[scale=0.41]{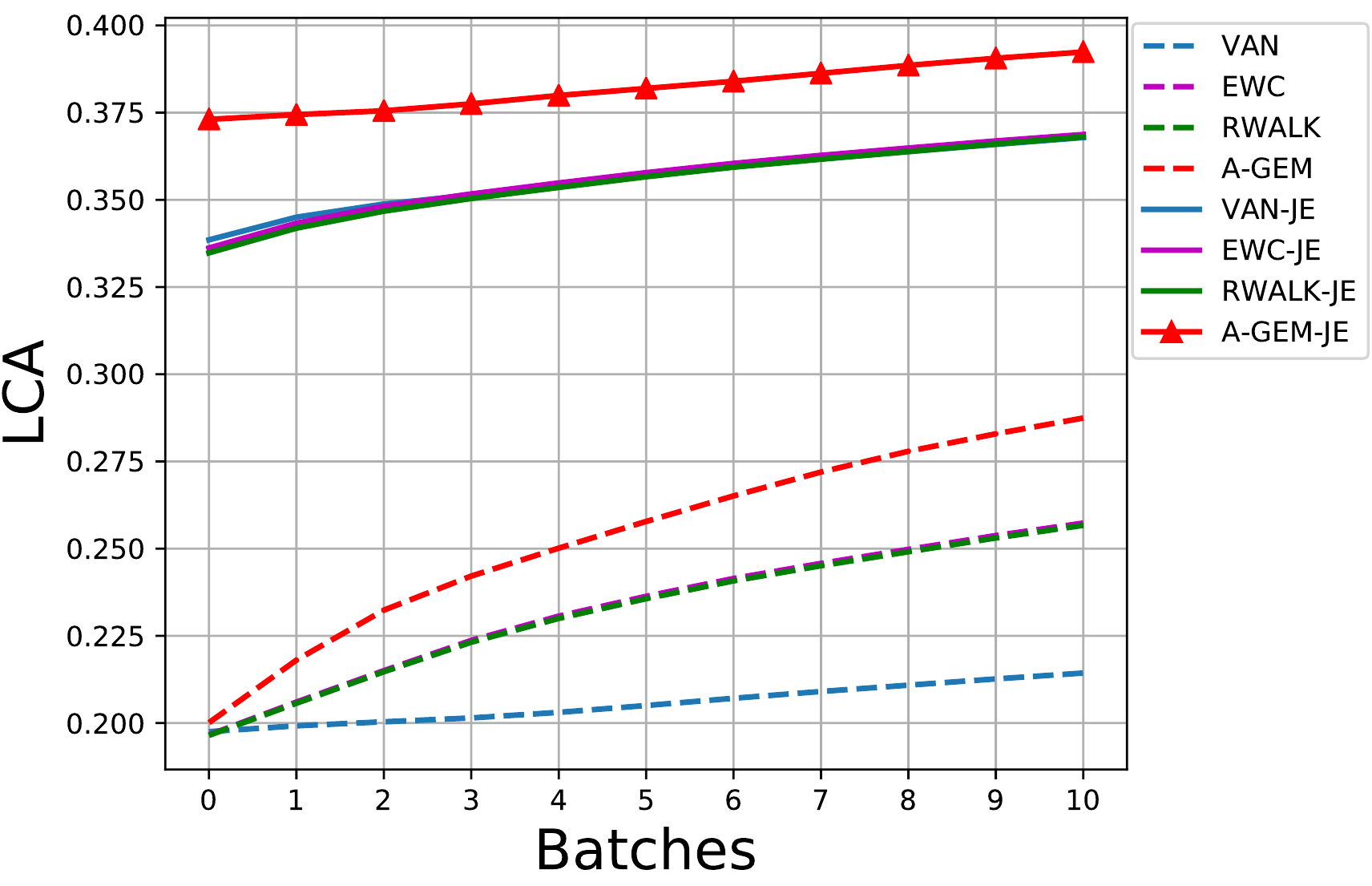}
                \caption{\small Split AWA}
        \end{center}
        \end{subfigure}%
\caption{\em Evolution of \famm during the first ten mini-batches.}
\label{fig:cub_awa_avg_acc_evol}
\end{figure}

Fig.~\ref{fig:grouped_cifar_mnist} and \ref{fig:grouped_cub_awa} show the overall results on all the datasets we considered 
(for brevity we show only representative methods, see detailed results in Appendix Tab.~\ref{tab:mnist_cifar_comp}, \ref{tab:cub_comp}, \ref{tab:awa_comp} and \ref{tab:gem_vs_sgem_time}). 
First, we observe that \sgem achieves the best average
accuracy on all datasets, except Permuted MNIST, where \pnn works better. The reason is because on this dataset each task has a large number of training examples,
which enables \pnn to learn its task specific parameters and to leverage its lateral connections. However, notice how \pnn has the worst memory cost by the end of training -
as its number of parameters grows super-linearly with the number of tasks. In particular, in large scale setups (Split CUB and AWA), \pnn runs out of memory during training due to its large size.
Also, \pnn does not learn well on datasets where tasks have fewer training examples.
Second, \sgem and \gem perform comparably in terms of average accuracy, but \sgem has much lower time (about $100$ times faster) and memory cost 
(about $10$ times lower), comparable to regularization-based approaches like EWC. 
Third, EWC and similar methods perform only slightly better than \van on this single pass LLL setting. The analysis in Appendix 
Sec.~\ref{sec:supp_increasing_epochs_and_arch} demonstrates that EWC  requires several epochs and over-parameterized architectures in order to work well.
Fourth, \pnn has no forgetting by construction and \sgem and \gem have the lowest forgetting among methods that use a fixed capacity architecture.
Next, all methods perform similarly in terms of \famm, with \pnn being the worst because of its ever growing number of parameters and \sgem slightly better than all
the other approaches. And finally, the use of task descriptors improves average accuracy across the board as shown in Fig.\ref{fig:grouped_cub_awa},
with \sgem a bit better than all the other methods we tried. All joint-embedding models using task descriptors have better \famm performance, although
this is the same across all methods including \sgem. Overall, we conclude that \sgem offers the best trade-off between average accuracy performance and 
efficiency in terms of sample, memory and computational cost.

Fig.~\ref{fig:mnist_cifar_avg_acc_evol} shows a more fine-grained analysis and comparison with more methods on Permuted MNIST and Split CIFAR.
The average accuracy plots show how \sgem and \gem greatly outperform other approaches, with the exception of \pnn on MNIST as discussed above.
On different datasets, different methods are best in terms of \famm, although \sgem is always top-performing. Fig.~\ref{fig:cub_awa_avg_acc_evol}
shows in more detail the gain brought by task descriptors which greatly speed up learning in the few-shot regime. On these datasets, \sgem performs the best or
on par to the best. 

\begin{figure}[t]
	\begin{subfigure}{0.45\linewidth}
        \begin{center}
                \includegraphics[scale=0.41]{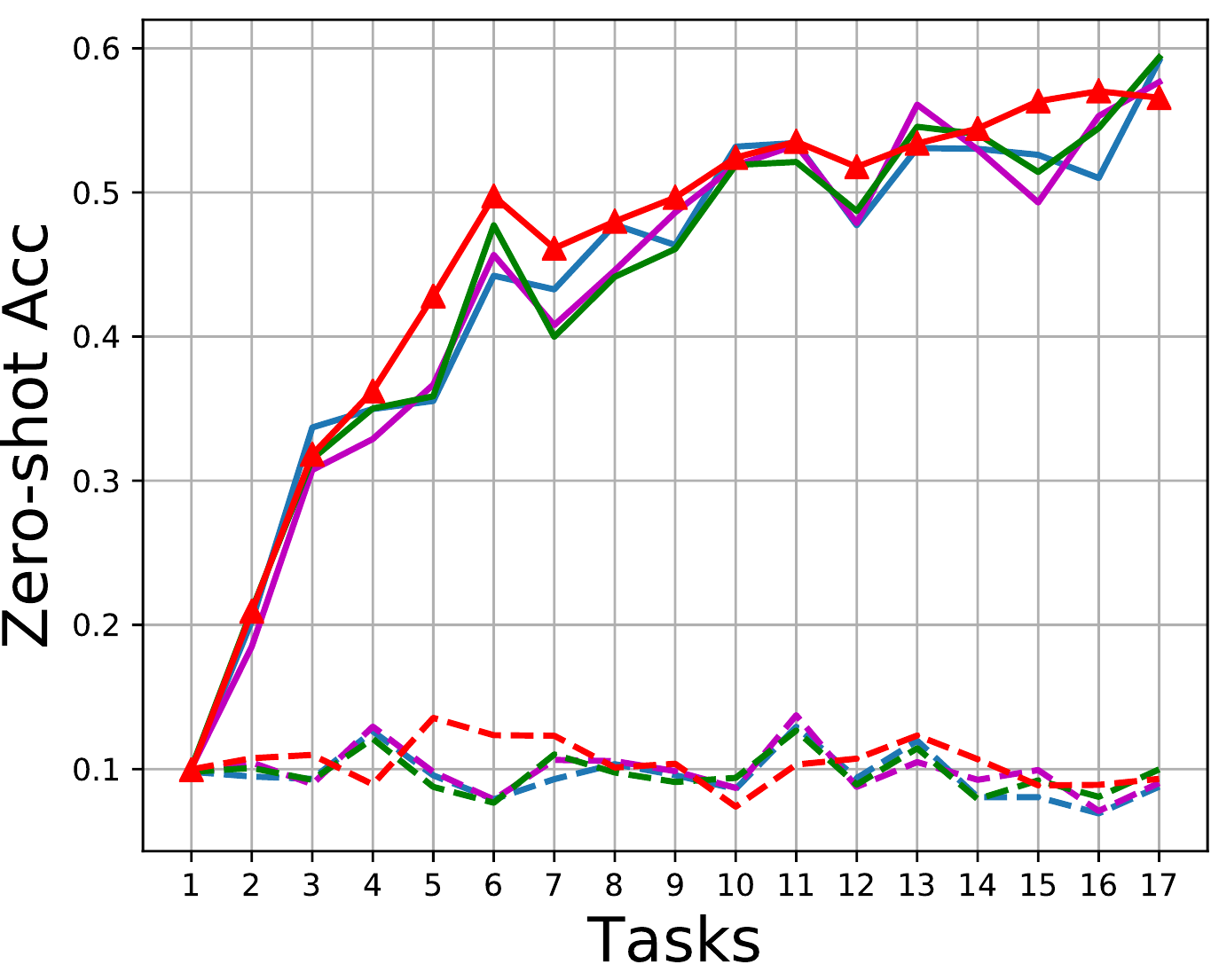}
                \caption{\small Split CUB}
        \end{center}
        \end{subfigure}%
        \begin{subfigure}{0.55\linewidth}
        \begin{center}
                \includegraphics[scale=0.41]{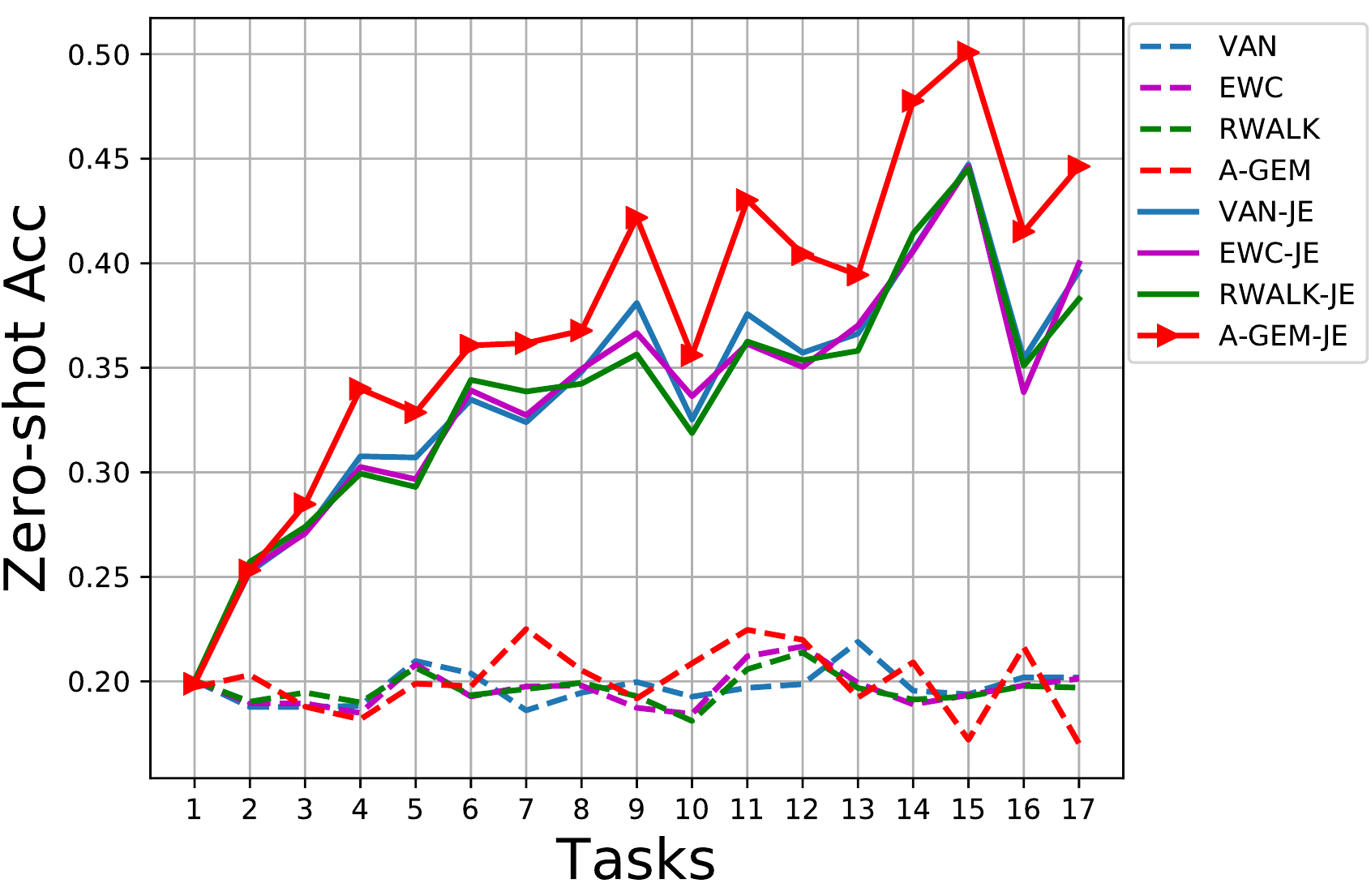}
                \caption{\small Split AWA}
        \end{center}
        \end{subfigure}%
\caption{\em Evolution of zero-shot performance as the learner sees new tasks on Split CUB and Split AWA datasets. }
\label{fig:awa20_zst_perf_evol}
\end{figure}

Finally, in Fig.~\ref{fig:awa20_zst_perf_evol}, we report the 0-shot performance of LLL methods on Split CUB and Split AWA datasets over time, showing a clear advantage of using compositional task descriptors with joint embedding models, which is more significant for \sgem. Interestingly, the zero-shot learning performance of joint embedding models  improves over time, indicating that these models get better at forward transfer or, in other words, become more \emph{efficient} over time. 

\SKIP{
\begin{figure}[tb]
        \begin{subfigure}{0.5\linewidth}
        \begin{center}
                \includegraphics[scale=0.38]{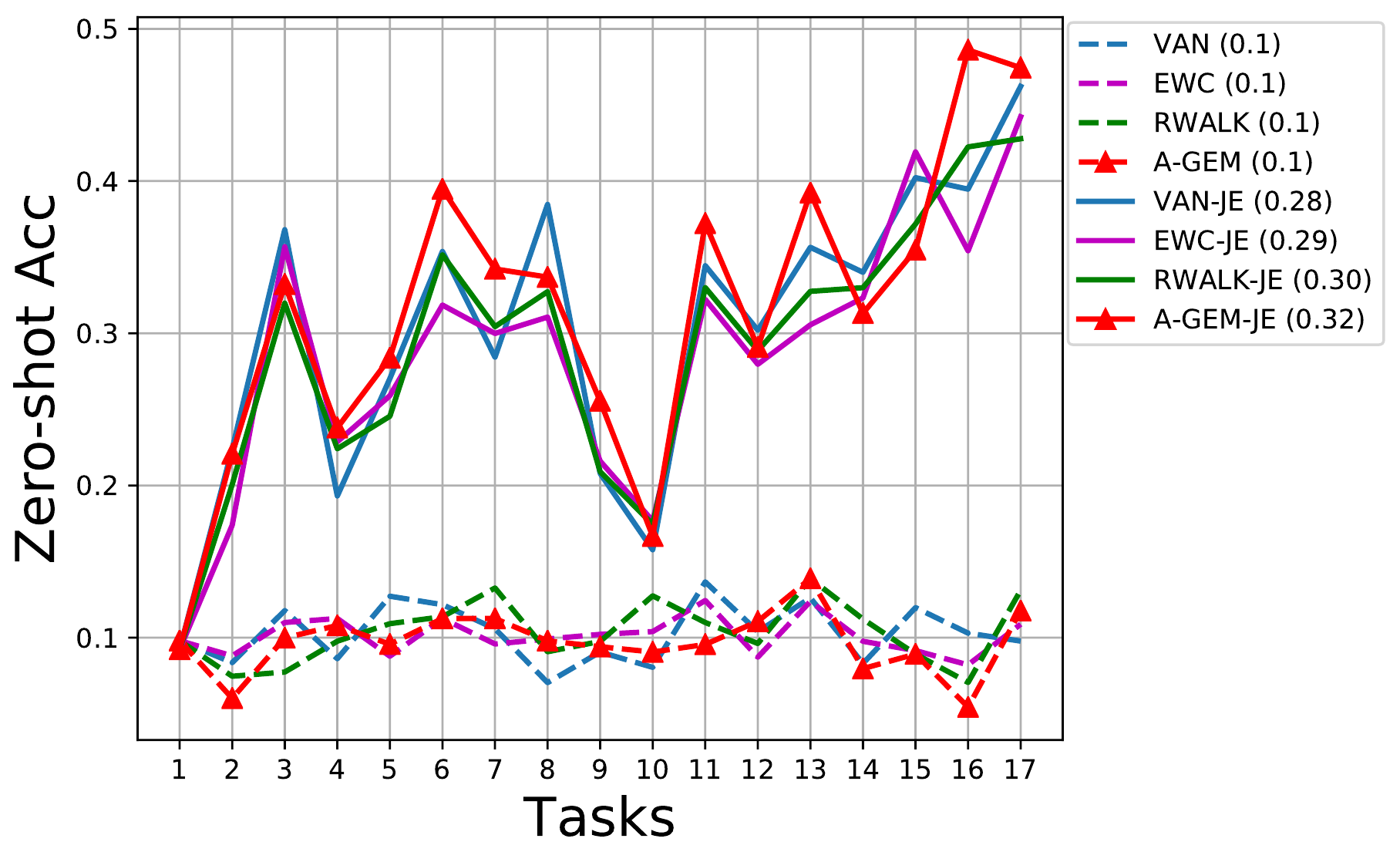}
                \caption{\small Split CUB}
        \end{center}
        \end{subfigure}%
        \begin{subfigure}{0.5\linewidth}
        \begin{center}
                \includegraphics[scale=0.38]{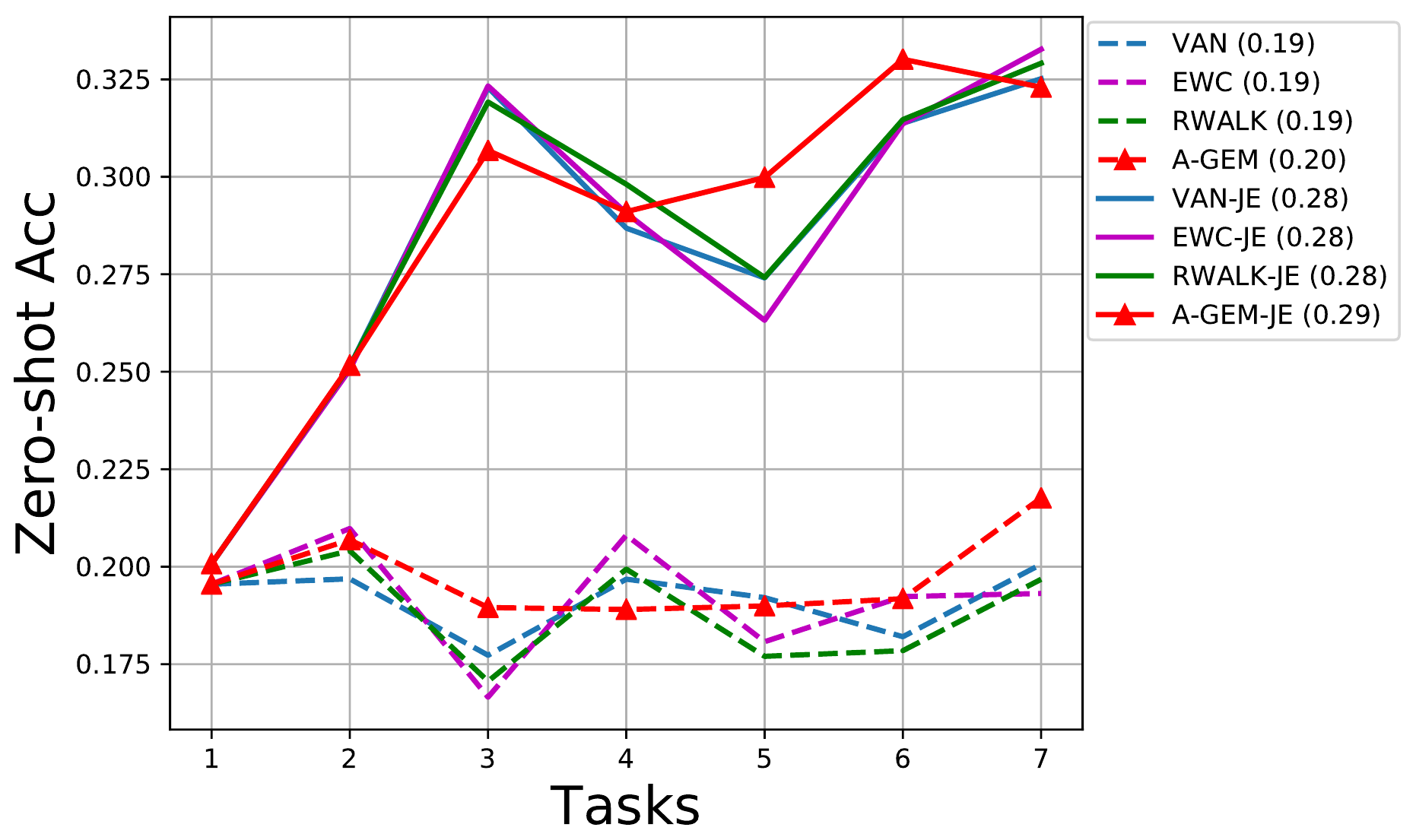}
                \caption{\small Split AWA}
        \end{center}
        \end{subfigure}
        
 
\caption{\em Evolution of zero-shot performance as the learner sees new tasks on  Split CUB and Split AWA datasets.  The average zero-shot performance across all tasks is shown between parenthesis after the name of each method.  }
\label{fig:zst_perf_evol}
\end{figure}}


\section{Related Work} \label{sec:related_work}
Continual~\citep{ring1997child} or Lifelong Learning (LLL)~\citep{thrun1998lifelong} have been the subject of extensive study over the past two decades. 
One approach to LLL uses modular compositional models~\citep{fernando2017pathnet, aljundi2017expert, rosenbaum2018routing, chang2018, rcl2018, modularmetal2018}, which limit interference among tasks by using different subset of modules for each task.
Unfortunately, these methods require searching over the space of architectures  
which is not sample efficient with current methods. 
Another approach is to regularize parameters important to solve past tasks~\citep{Kirkpatrick2016EWC, Zenke2017Continual, chaudhry2018riemannian}, which has been proven effective for over-parameterized models in the multiple epoch setting 
(see Appendix  Sec.~\ref{sec:supp_increasing_epochs_and_arch}),
while we focus on learning from few examples using memory efficient models. 
Methods based on episodic memory~\citep{Rebuffi16icarl, lopez2017gradient} require a little bit more memory at training time but can 
work much better in the single pass setting we considered~\citep{lopez2017gradient}.

The use of task descriptors for LLL has already been advocated by
\citet{Isele:2016} but using a sparse coding framework which is not obviously 
applicable to deep nets in a computationally efficient way, and
also by \citet{lopez2017gradient} although they did not explore the use of compositional
descriptors. More generally, tasks descriptors have been used in Reinforcement Learning
 with similar motivations by several others~\citep{horde, uvfa, commai2}, and it is also
a key ingredient in all the zero/few-shot learning algorithms~\citep{lampert2014attribute,xian2018zero,elhoseiny2017write,WahCUB_200_2011,lampert2009learning}.

\section{Conclusion} \label{sec:conclusion}

We studied the problem of efficient Lifelong Learning (LLL) in the case where the learner can only do a single pass over the input data stream.
We found that our approach, \sgem, has the best trade-off between average accuracy by the end of the learning experience and computational/memory cost.
Compared to the original \gem\ algorithm, \sgem\ is about 100 times faster and has 10 times less memory requirements; compared to regularization based approaches,
it achieves significantly higher average accuracy. We also demonstrated that by
using compositional task descriptors all methods can improve their few-shot performance, with \sgem often being the best.


Our detailed experiments reported in Appendix~\ref{sec:detailedresults} also show that there is still a substantial performance gap 
between LLL methods, including \sgem, trained in a sequential learning setting and the same network trained in a non-sequential multi-task setting, despite seeing the same data samples. Moreover, while task descriptors do help in the few-shot learning regime, the \famm performance gap between different methods is very small; suggesting a poor ability of current methods to transfer knowledge even when forgetting has been eliminated. Addressing these two fundamental issues will be the focus of our future research.

\clearpage

\bibliographystyle{iclr2019_conference}
\bibliography{chaudhryBibliography}

\clearpage

\section*{Appendix}
In Sec.~\ref{sec:dataset_stats} we report the summary of datasets used for the experiments. Sec.~\ref{sec:agem:algorithm} details our \sgem algorithm and Sec.~\ref{sec:supp_sgem_update_rule} provides the proof of update rule of \sgem discussed in Sec.~\ref{sec:approach} of the main paper. In Sec.~\ref{sec:gem_vs_sgem_abal}, we analyze the differences between \sgem and \gem, and describe another variation of \gem, dubbed Stochastic \gem (\mgem). 
The detailed results of the experiments which were used to generate Fig~\ref{fig:grouped_cifar_mnist} and \ref{fig:grouped_cub_awa} in the main paper are given in Sec.~\ref{sec:detailedresults}. In Sec.~\ref{sec:supp_increasing_epochs_and_arch}, we provide empirical evidence to the conjecture that regularization-based approaches like \ewc require over-parameterized architectures and multiple passes over data in order to perform well as discussed in the Sec.~\ref{subsec:results} of the main paper. In Sec.~\ref{sec:supp_hyperparam_selec}, we provide the grid used for the cross-validation of different hyper-parameters and report the optimal values for different models. Finally, in Sec.~\ref{sec:je_pict_desc}, we pictorially describe the joint embedding model discussed in Sec.~\ref{sec:joint_embed}. 

\appendix

\section{Dataset Statistics}
\label{sec:dataset_stats}
\begin{table}[bh]
\small
\centering
\caption{\em \textbf{Dataset statistics}.}
\label{tab:datasets}
\begin{tabular}{l|cccc}
\toprule
  & \textbf{Perm. MNIST} &\textbf{Split CIFAR} & \textbf{Split CUB} & \textbf{Split AWA} \\
\hline
num. of tasks & 20 & 20 & 20 & 20 \\
input size & 1$\times$28$\times$28 & 3$\times$32$\times$32 & 3$\times$224$\times$224 & 3$\times$224$\times$224 \\ 
num. of classes per task & 10 & 5 & 10 & 5 \\
num. of training images per task & 60000 & 2500 & 300 & - \\ 
num. of test images per task & 10000 & 500 & 290 & 560 \\  
\bottomrule
\end{tabular}
\end{table}

\section{\sgem Algorithm}
\label{sec:agem:algorithm}
\begin{algorithm}[h!]
\caption{Training and evaluation of \sgem\ on sequential data $\dataset = \{\dataset_1, \cdots, \dataset_T\}$} \label{alg:sgem}
\footnotesize
\begin{multicols}{2}
\begin{algorithmic}[1]
  \Procedure{TRAIN}{$f_{\param}, \dataset^{train}, \dataset^{test}$}
    \State $\epsmem\gets \{\}$
    \State $A \gets 0 \in \real^{T \times T}$
    \For{$t=\{1,\cdots,T\}$}
                \For{$(\ip,y) \in \dataset_t^{train}$}
                        \State $(\ip_{ref}, y_{ref})\sim\epsmem$
                        \State $g_{ref}\gets\nabla_{\param} \ell(f_{\theta}(\ip_{ref},t), y_{ref})$
                        \State $g\gets\nabla_{\param} \ell(f_{\theta}(\ip,t), y)$
                                                      \If{$g^\top g_{ref} \geq 0$}
                                                                               \State $\tilde{g}\gets g$
                                                                                                      \Else
                                                                                                      \State $\tilde{g}\gets g - \frac{g^\top g_{ref}}{g_{ref}^\top g_{ref}}g_{ref}$
                        \EndIf
                        \State $\param \gets \param - \alpha\tilde{g}$
                \EndFor
                \State $\epsmem \gets \textproc{UpdateEpsMem}(\epsmem, \dataset_t^{train}, T)$
                \State $A_{t,:} \gets \textproc{EVAL}(f_{\param}, \dataset^{test})$
        \EndFor
    \State \textbf{return} $f_{\param}, A$
  \EndProcedure
\end{algorithmic}
\columnbreak
\begin{algorithmic}[1]
   \Procedure{EVAL}{$f_{\param}, \dataset^{test}$}
                \State $a \gets 0 \in \real^T$
       \For{$t=\{1,\cdots,T\}$}
                \State $a_t \gets 0$
            \For{$(\ip,y) \in \dataset_t^{test}$}
                          \State $a_t \gets a_t + \textproc{accuracy}(f_{\theta}(\ip, t), y)$
            \EndFor
            \State $a_t \gets \frac{a_t}{len(\dataset_t^{test})}$
       \EndFor
       \State \textbf{return} $a$
   \EndProcedure
\end{algorithmic}
\vspace{2mm}
\begin{algorithmic}[1]
   \Procedure{UpdateEpsMem}{$\epsmem, \dataset_t, T$}
                \State $s \gets \frac{|\epsmem|}{T}$
        \For{$i=\{1,\cdots,s\}$}
                \State $(\ip,y) \sim \dataset_t$
            \State $\epsmem \gets (\ip,y)$
        \EndFor
        \State \textbf{return} $\epsmem$
   \EndProcedure
\end{algorithmic}
\end{multicols}
\end{algorithm}

\section{\sgem Update Rule}
\label{sec:supp_sgem_update_rule}

Here we provide the proof of the update rule of \sgem (Eq.~\ref{eq:agem_update_rule}), $\tilde{g} = g - \frac{g^\top g_{ref}}{g_{ref}^\top g_{ref}}g_{ref}$, stated in Sec.~\ref{sec:approach} of the main paper.
\begin{proof}
The optimization objective of \sgem\, as described in the Eq.~\ref{eq:s_gem_primal_obj} of the main paper, is:
\begin{align*}
	 	\textrm{minimize}_{\tilde{g}} \quad &\frac{1}{2}||g-\tilde{g}||_2^2 \\
	 	\textrm{s.t.} 				\quad &\tilde{g}^\top g_{ref} \geq 0 \numberthis \label{eq:supp_sgem_obj}  
\end{align*}
Replacing $\tilde{g}$ with $z$ and rewriting Eq.~\ref{eq:supp_sgem_obj} yields:
\begin{align*}
	 	\textrm{minimize}_{z} \quad &\frac{1}{2}z^\top z - g^\top z \\
	 	\textrm{s.t.} 				\quad &- z^\top g_{ref} \leq 0  \numberthis \label{eq:supp_sgem_simp_obj} 
\end{align*}
Note that we discard the term $g^\top g$ from the objective and change the sign of the inequality constraint. The Lagrangian of the constrained optimization problem defined above can be written as:
\begin{equation}
	\label{eq:supp_sgem_lagrang}
	\mathcal{L}(z, \alpha) = \frac{1}{2}z^\top z - g^\top z - \alpha z^\top g_{ref}
\end{equation}

Now, we pose the dual of Eq.~\ref{eq:supp_sgem_lagrang} as:
\begin{equation}
	\label{eq:supp_sgem_dual}
	 \theta_{\mathcal{D}}(\alpha) =  \min_{z} \mathcal{L}(z, \alpha)
\end{equation}

Lets find the value $z^*$ that minimizes the $\mathcal{L}(z, \alpha)$ by setting the derivatives of $\mathcal{L}(z, \alpha)$ \wrt to $z$ to zero:
\begin{align*}
	\nabla_{z} \mathcal{L}(z, \alpha) = 0 \\
	z^* =   g + \alpha g_{ref} \numberthis \label{eq:supp_sgem_z_star}
\end{align*}

The simplified dual after putting the value of $z^*$ in Eq.~\ref{eq:supp_sgem_dual} can be written as:
\begin{align*}
	 \theta_{\mathcal{D}}(\alpha) & = \frac{1}{2}(g^\top g + 2\alpha g^\top g_{ref} + \alpha^2 g_{ref}^\top g_{ref}) - g^\top g -2\alpha g^\top g_{ref} - \alpha^2 g_{ref}^\top g_{ref} \\
	& =  -\frac{1}{2}g^\top g - \alpha g^\top g_{ref} - \frac{1}{2} \alpha^2 g_{ref}^\top g_{ref}
\end{align*}

The solution $\alpha^* = \max_{\alpha; \alpha>0} \theta_{\mathcal{D}}(\alpha)$ to the dual is given by:
\begin{align*}
	\nabla_{\alpha} \theta_{\mathcal{D}}(\alpha) = 0 \\
	\alpha^* = - \frac{g^\top g_{ref}}{g_{ref}^\top g_{ref}}
\end{align*}

By putting $\alpha^*$ in Eq.~\ref{eq:supp_sgem_z_star}, we recover the \sgem\ update rule:
\begin{equation*}
	z^* = g  - \frac{g^\top g_{ref}}{g_{ref}^\top g_{ref}} g_{ref} = \tilde{g}
\end{equation*}
\end{proof}

\section{Analysis of gem\ and \sgem} \label{sec:gem_vs_sgem_abal}

\begin{figure}[tb]
	\begin{subfigure}{0.5\linewidth}
	\begin{center}
		\includegraphics[scale=0.4]{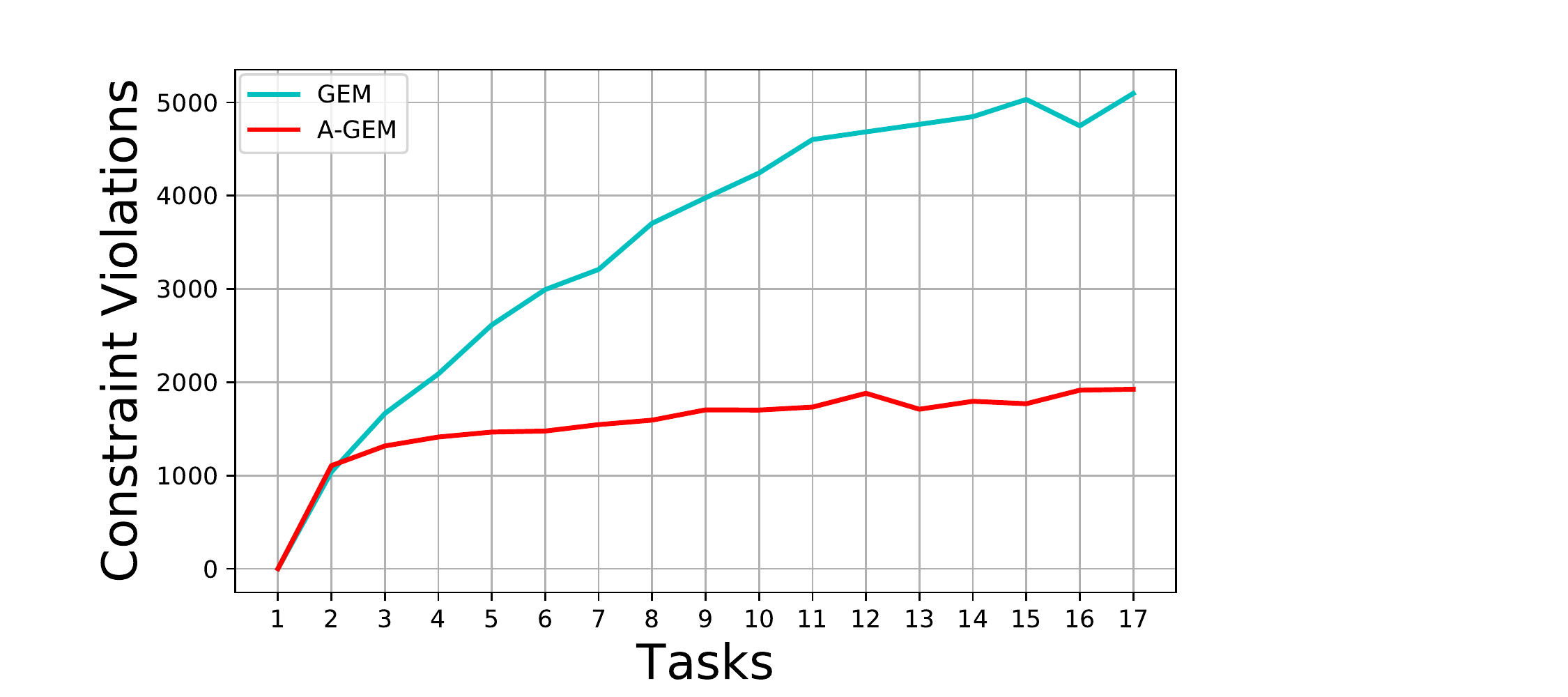}
		\caption{\small MNIST}
	\end{center}
	\end{subfigure}%
	\begin{subfigure}{0.5\linewidth}
	\begin{center}
		\includegraphics[scale=0.4]{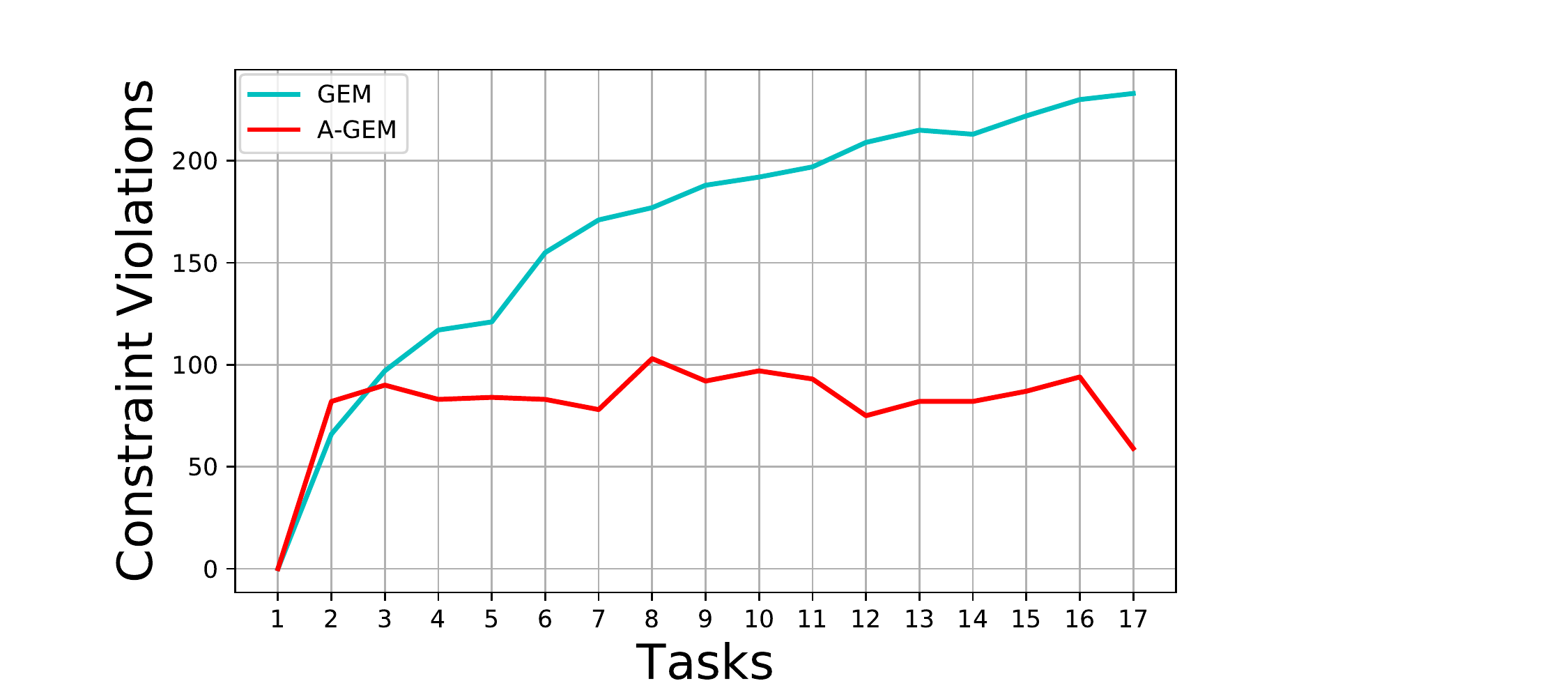}
		\caption{\small CIFAR}
	\end{center}
	\end{subfigure}%
	\vspace{-3mm}
\caption{\em Number of constraint violations in \textbf{\gem} and \textbf{\sgem} on \textbf{Permuted MNIST} and \textbf{Split CIFAR} as new tasks are learned.}
	\label{fig:gem_vs_sgem_constrnt_violtn}
\end{figure}

In this section, we empirically analyze the differences between \sgem\ and \gem, and report experiments with 
another computationally efficient but worse performing version of \gem.
\subsection{Frequency of Constraint Violations}
\label{sec:supp_freq_const_violtn}
Fig.~\ref{fig:gem_vs_sgem_constrnt_violtn} shows the frequency of constraint violations (see Eq.~\ref{eq:gem_dual_obj} and \ref{eq:s_gem_primal_obj}) 
on Permuted MNIST and Split CIFAR datasets. 
Note that, the number of gradient updates (training steps) per task on MNIST and CIFAR are $5500$ and $250$, respectively. 
As the number of tasks increase, \gem\ violates the optimization constraints at almost each training step, whereas \sgem\ plateaus to a much lower value. 
Therefore, the computational efficiency of \sgem\ not only stems from the fact that it avoids solving a QP at each training step (which is much more expensive than a simple
inner product) but also from the fewer number of constraint violations. From the figure, we can also infer that as the number of tasks grows the gap between \gem\ 
and \sgem\ would grow further. Thus, the computational and memory overhead of \gem\ over \sgem, see also Tab.~\ref{tab:gem_vs_sgem_time}, 
gets worse as the number of tasks increases.

\subsection{Average Accuracy and Worst-Case Forgetting}
\label{sec:avg_Acc_wst_fgt}

In Tab.~\ref{tab:avg_acc_wst_fgt}, 
we empirically demonstrate the different properties induced by the objective functions of \gem\ and \sgem. \gem\ enjoys lower worst-case task forgetting
while \sgem\ enjoys better overall average accuracy. This is particularly true on the training examples stored in memory, as on the test set the result is confounded
by the generalization error.

\begin{table}[bh]
\centering
\caption{\em Comparison of average accuracy ($A_T$) and worst-case forgetting ($F_{wst}$) on the \textbf{Episodic Memory ($\epsmem$)} and \textbf{Test Set ($\dataset^{EV}$)}.}
\label{tab:avg_acc_wst_fgt}
\begin{tabular}{@{}c@{\hspace{14pt}}c@{\hspace{4pt}}c@{\hspace{10pt}}c@{\hspace{4pt}}c@{\hspace{12pt}}c@{\hspace{4pt}}c@{\hspace{10pt}}c@{\hspace{4pt}}c@{}}
\toprule
\multicolumn{1}{c}{\textbf{Methods}} & \multicolumn{4}{c}{\textbf{MNIST}}                                             & \multicolumn{4}{c}{\textbf{CIFAR}}                                             \\ 
\midrule
        & \multicolumn{2}{c}{$\epsmem$} & \multicolumn{2}{c}{$\dataset^{EV}$} & \multicolumn{2}{c}{$\epsmem$} & \multicolumn{2}{c}{$\dataset^{EV}$} \\  \cmidrule(r){2-5} \cmidrule(l){6-9}
        & \multicolumn{1}{c}{$A_{T}$}                & \multicolumn{1}{c}{$F_{wst}$}               & \multicolumn{1}{c}{$A_{T}$}               & \multicolumn{1}{c}{$F_{wst}$}              & \multicolumn{1}{c}{$A_{T}$}                & \multicolumn{1}{c}{$F_{wst}$}               & \multicolumn{1}{c}{$A_{T}$}               & \multicolumn{1}{c}{$F_{wst}$}              \\
        \cmidrule(r){2-3} \cmidrule(r){4-5} \cmidrule(l){6-7} \cmidrule(l){8-9}
\gem     & \multicolumn{1}{c}{99.5}                 & \multicolumn{1}{c}{0}                & \multicolumn{1}{c}{89.5}                &  \multicolumn{1}{c}{0.10}              & \multicolumn{1}{c}{97.1}                 & \multicolumn{1}{c}{0.05}                & \multicolumn{1}{c}{61.2}                & \multicolumn{1}{c}{0.14}                \\
\sgem   & \multicolumn{1}{c}{99.3}                 & \multicolumn{1}{c}{0.008}                & \multicolumn{1}{c}{89.1}                 & \multicolumn{1}{c}{0.13}               &  \multicolumn{1}{c}{72.1}                &  \multicolumn{1}{c}{0.15}               & \multicolumn{1}{c}{62.3}                & \multicolumn{1}{c}{0.15}                \\ 
\bottomrule
\end{tabular}
\end{table}

\subsection{Stochastic \gem (\mgem)}
\label{sec:supp_stochstc_gem}
In this section we report experiments with another variant of \gem, dubbed Stochastic \gem\ (\mgem). The main idea in \mgem\ is to randomly sample one constraint, 
at each training step, from the possible $t-1$ constraints of \gem. If that constraint is violated, the gradient is projected only taking into account that 
constraint. Formally, the optimization objective of \mgem is given by:
\begin{align*}
	 \textrm{minimize}_{\tilde{g}} \quad &\frac{1}{2}||g-\tilde{g}||_2^2 \\
	 \textrm{s.t.} 				\quad &\langle\tilde{g}, g_k\rangle \geq 0 \quad \textrm{where } k \sim \{1,\cdots,t-1\} \numberthis \label{eq:mgem_primal_obj}
\end{align*} 

In other words, at each training step, \mgem\ avoids the increase in loss of one of the previous tasks sampled randomly. 
In Tab.~\ref{tab:agem_sgem_comp} we report the comparison of \gem, \mgem\ and \sgem\ on Permuted MNIST and Split CIFAR. 

Although, \mgem is closer in spirit to \gem, as it requires randomly sampling one of the \gem constraints to satisfy, compared to \sgem, which defines the constraint as the average gradient of the previous tasks, it perform slightly worse than \gem, as can be seen from Tab.~\ref{tab:agem_sgem_comp}.

\begin{table}[H]
\centering
\small
\caption{\em Comparison of different variations of \gem on MNIST Permutations and Split CIFAR.}
\label{tab:agem_sgem_comp}
\begin{tabular}{lcc|cc}
\toprule
\multicolumn{1}{l}{\textbf{Methods}} &\multicolumn{2}{c}{\textbf{Permuted MNIST}} &\multicolumn{2}{c}{\textbf{Split CIFAR}} \\
\hline
                              & $A_{T}$(\%)        & $F_{T}$     & $A_{T}$(\%)      & $F_{T}$  \\
  \cmidrule(r){2-3} \cmidrule(l){4-5}
{\gem}                  & 89.5          & 0.06      & 61.2  &  0.06        \\
{\mgem} 		        & 88.2        & 0.08  		   & 56.2   &  0.12   			 \\
{\sgem}            	    & 89.1 	    & 0.06       	   & 62.3  & 0.07        \\
  \bottomrule
\end{tabular}
\end{table}

\section{Result Tables} \label{sec:detailedresults}
In Tab.~\ref{tab:mnist_cifar_comp},~\ref{tab:cub_comp},~\ref{tab:awa_comp} and~\ref{tab:gem_vs_sgem_time} we report the detailed results which were used to generate Fig.\ref{fig:grouped_cifar_mnist} and \ref{fig:grouped_cub_awa}.
\begin{table}[h]
\centering
\small
\caption{\em Comparison with different baselines on Permuted MNIST and Split CIFAR. The value of $\infty$ is assigned to a metric when the model fails to train with the cross-validated values of hyper-parameters found on the subset of the tasks as discussed in Sec.~\ref{sec:setup} of the main paper. The numbers are averaged across 5 runs using a different seed each time. The results from this table are used to generate Fig~\ref{fig:grouped_cifar_mnist} in Sec.~\ref{subsec:results} of the main paper.}
\label{tab:mnist_cifar_comp}
\resizebox{\textwidth}{!}{%
\begin{tabular}{lccc|ccc}
\toprule
\multicolumn{1}{l}{\textbf{Methods}} &\multicolumn{3}{c}{\textbf{Permuted MNIST}} &\multicolumn{3}{c}{\textbf{Split CIFAR}} \\
\hline
        & $A_{T}$(\%)  & $F_{T}$ & $LCA_{10}$  & $A_{T}$(\%) & $F_{T}$  & $LCA_{10}$\\
\cmidrule(r){2-4} \cmidrule(l){5-7}
{\van}                      & 47.9 (\textpm\ 1.32)      & 0.51 (\textpm\ 0.01) & 0.26 (\textpm\ 0.006)          & 42.9 (\textpm\ 2.07)  & 0.25 (\textpm\ 0.03) & 0.30 (\textpm\ 0.008)        \\
{\icarl}                    & -        & -  & -  & 50.1  & 0.11 & -                            \\
{\ewc}                      & 68.3 (\textpm\ 0.69)     & 0.29 (\textpm\ 0.01) & 0.27 (\textpm\ 0.003)          & 42.4 (\textpm\ 3.02 ) & 0.26 (\textpm\ 0.02) & 0.33 (\textpm\ 0.01)     \\
{\pii}                      & $\infty$  & $\infty$  & $\infty$    & 47.1 (\textpm\ 4.41)   & 0.17 (\textpm\ 0.04) & 0.31 (\textpm\ 0.008)    \\
{\mas}                      & 69.6 (\textpm\ 0.93)      & 0.27 (\textpm\ 0.01)    & 0.29 (\textpm\ 0.003)      & 44.2 (\textpm\ 2.39) & 0.25 (\textpm\ 0.02)  & 0.33 (\textpm\ 0.009)  \\
{\rwalk}                    & 85.7 (\textpm\ 0.56)       & 0.08 (\textpm\ 0.01)  & \textbf{0.31} (\textpm\ 0.005)        & 40.9 (\textpm\ 3.97)   & 0.29 (\textpm\ 0.04) & 0.32 (\textpm\ 0.005)   \\
{\pnn}                      & \textbf{93.5} (\textpm\ 0.07)  & \textbf{0}    & 0.19 (\textpm\ 0.006)         & 59.2 (\textpm\ 0.85) & \textbf{0}    & 0.21 (\textpm\ 0.001)    \\
{\gem}                      & 89.5 (\textpm\ 0.48)     & 0.06 (\textpm\ 0.004) & 0.23 (\textpm\ 0.005) & 61.2 (\textpm\ 0.78) & 0.06 (\textpm\ 0.007)& \textbf{0.36} (\textpm\ 0.007) \\
\textbf{\sgem (Ours)}       & 89.1 (\textpm\ 0.14) & 0.06 (\textpm\ 0.001) & 0.29 (\textpm\ 0.004) & \textbf{62.3} (\textpm\ 1.24)     & 0.07 (\textpm\ 0.01) & 0.35 (\textpm\ 0.01) \\
\cmidrule(r){2-4} \cmidrule(l){5-7}
{\mtask}                    & 95.3      & - & -             & 68.3  & - & - \\
\bottomrule
\end{tabular}}
\end{table}

\begin{table}[h]
\small
\centering
\caption{\em Average accuracy and forgetting of standard models (left) and joint embedding models (right) on Split CUB. The value of `OoM' is assigned to a metric when the model fails to fit in the memory. The numbers are averaged across 10 runs using a different seed each time. The results from this table are used to generate Fig~\ref{fig:grouped_cub_awa} in Sec.~\ref{subsec:results} of the main paper. }
\label{tab:cub_comp}
\resizebox{\textwidth}{!}{%
\begin{tabular}{lccc}
\toprule
\textbf{Methods} & \multicolumn{3}{c}{\textbf{Split CUB}}  \\
\cmidrule{1-4}
            & $A_{T}$(\%)        & $F_{T}$   & $LCA_{10}$  \\
\cmidrule{2-4}
{\van}                        & 54.3 (\textpm\ 2.03) /\ 67.1 (\textpm\ 4.77)                   & 0.13 (\textpm\ 0.02)/\ 0.10 (\textpm\ 0.04) & 0.29 (\textpm\ 0.009) /\ 0.52 (\textpm\ 0.01) \\
{\ewc}                        & 54 (\textpm\ 1.08) /\ 68.4 (\textpm\ 4.08)                   & 0.13 (\textpm\ 0.02) /\ 0.09 (\textpm\ 0.03) & 0.29 (\textpm\ 0.007) /\ 0.52 (\textpm\ 0.01) \\
{\pii}                        & 55.3 (\textpm\ 2.28) /\ 66.6 (\textpm\ 5.18)                   & 0.12 (\textpm\ 0.02)/\ 0.10 (\textpm\ 0.04)  & 0.29 (\textpm\ 0.008) /\ 0.52 (\textpm\ 0.01) \\
{\rwalk}                      & 54.4 (\textpm\ 1.82) /\ 67.4 (\textpm\ 3.50)                   & 0.13 (\textpm\ 0.01) /\ 0.10 (\textpm\ 0.03)  & 0.29 (\textpm\ 0.008) /\ 0.52 (\textpm\ 0.01) \\
{\pnn}                        & OoM /\ OoM  & OoM /\ OoM & OoM /\ OoM \\
\textbf{\sgem (Ours)}         & \textbf{62} (\textpm\ 3.5) /\ \textbf{71} (\textpm\ 2.83)  & \textbf{0.07} (\textpm\ 0.02) /\ \textbf{0.07} (\textpm\ 0.01)  & \textbf{0.30} (\textpm\ 0.008) /\ \textbf{0.54} (\textpm\ 0.02) \\
\cmidrule{2-4}
{\mtask}                      & 65.6 /\ 73.8 & - /\ - & - /\ - \\
\bottomrule
\end{tabular}}
\end{table}

\begin{table}[h]
\small
\centering
\caption{\em Average accuracy and forgetting of standard models (left) and joint embedding models (right) on Split AWA. The value of `OoM' is assigned to a metric when the model fails to fit in the memory. The numbers are averaged across 10 runs using a different seed each time. The results from this table are used to generate Fig~\ref{fig:grouped_cub_awa} in Sec.~\ref{subsec:results} of the main paper. }
\label{tab:awa_comp}
\resizebox{\textwidth}{!}{%
\begin{tabular}{lccc}
\toprule
\textbf{Methods} & \multicolumn{3}{c}{\textbf{Split AWA}}  \\
\cmidrule{1-4}
            & $A_{T}$(\%)        & $F_{T}$   & $LCA_{10}$  \\
\cmidrule{2-4}
{\van}  & 30.3 (\textpm\ 2.84) /\ 42.8 (\textpm\ 2.86)         & \textbf{0.04} (\textpm\ 0.01) /\ 0.07 (\textpm\ 0.02) & 0.21 (\textpm\ 0.008) /\ 0.37 (\textpm\ 0.02) \\
{\ewc}  & 33.9 (\textpm\ 2.87) /\ 43.3 (\textpm\ 3.71)         & 0.08 (\textpm\ 0.02) /\ 0.07 (\textpm\ 0.03) & 0.26 (\textpm\ 0.01) /\ 0.37 (\textpm\ 0.02) \\
{\pii}  & 33.9 (\textpm\ 3.25) /\ 43.4 (\textpm\ 3.49)         & 0.08 (\textpm\ 0.02) /\ 0.06 (\textpm\ 0.02) & 0.26 (\textpm\ 0.01) /\ 0.37 (\textpm\ 0.02)  \\
{\rwalk} & 33.9 (\textpm\ 2.91) /\ 42.9 (\textpm\ 3.10)         & 0.08 (\textpm\ 0.02) /\ 0.07 (\textpm\ 0.02) & 0.26 (\textpm\ 0.01) /\ 0.37 (\textpm\ 0.02)  \\
{\pnn} & OoM /\  OoM & OoM /\  OoM & OoM /\  OoM \\
\textbf{\sgem (Ours)} & \textbf{44} (\textpm\ 4.10) /\ \textbf{50} (\textpm\ 3.25)  &
0.05 (\textpm\ 0.02) /\ \textbf{0.03} (\textpm\ 0.02) & \textbf{0.29} (\textpm\ 0.01) /\ \textbf{0.39} (\textpm\ 0.02) \\
\cmidrule{2-4}
{\mtask} & 64.8 /\ 66.8         & - /\ - & - /\ -  \\
\bottomrule
\end{tabular}}
\end{table}

\begin{table}[H]
\small
\centering
\caption{\em Computational cost and memory complexity of different LLL approaches. The timing refers to training time on a GPU device.
Memory cost is provided in terms of the total number of parameters P, the size of the minibatch B, the total size of the network hidden state H (assuming
all methods use the same architecture), the size of the episodic memory M per task. The results from this table are used to generate Fig.~\ref{fig:grouped_cifar_mnist} and \ref{fig:grouped_cub_awa} in Sec.~\ref{subsec:results} of the main paper. 
}
\label{tab:gem_vs_sgem_time}
\begin{tabular}{l cccc || cc} 
\toprule
\textbf{Methods} & \multicolumn{4}{c}{\textbf{Training Time [s]}} & \multicolumn{2}{c}{\textbf{Memory}} \\ \hline
        & MNIST         & CIFAR  & CUB & AWA    & Training & Testing         \\
\cmidrule(r){2-5} \cmidrule(l){6-7}
\van     & 186          & 105 & 54 & 4123 & P + B*H  & P + B*H               \\
\ewc     & 403          & 250 & 72 & 4136 & 4*P + B*H & P + B*H               \\
\pn      & 510 & 409          & $\infty$ & $\infty$ & 2*P*T + B*H*T & 2*P*T + B*H*T  \\
\gem     & 3442    & 5238     & - & - & P*T + (B+M)*H & P + B*H              \\
\textbf{\sgem (Ours)}  & 477  & 449 & 420 & 5221  & 2*P + (B+M)*H & P + B*H \\
\bottomrule
\end{tabular}
\end{table}

\section{Analysis of EWC}
\label{sec:supp_increasing_epochs_and_arch}
In this section we provide empirical evidence to the conjecture that regularization-based approaches like \ewc need over-parameterized architectures and
multiple passes over the samples of each task in order to perform well. The intuition as to why models need to be over-parameterized is 
because it is easier to avoid cross-task interference when the model has additional capacity. 
In the single-pass setting and when each task does not have very many training samples, 
regularization-based appraches also suffer because regularization parameters cannot be estimated well from a model 
that has not fully converged. Moreover, for tasks that do not have much data, rgularization-based
approaches do not enable any kind of positive backward transfer~\citep{lopez2017gradient} which further hurts performance as the predictor cannot leverage knowledge
acquired later to improve its prediction on past tasks. Finally, regularization-based approaches perform much better in the multi-epoch setting simply because
in this setting the baseline un-regularized model performs much worse, as it overfits much more to the data of the current task, every time unlearning what it 
learned before.

We consider Permuted MNIST and Split CIFAR datasets as described in Sec.~\ref{sec:experiments} of the main paper. 
For MNIST, the two architecture variants that we experiment with are; $1$) two-layer fully-connected network with $256$ units in each layer (denoted by $-S$ suffix), and $2$) 
two-layer fully-connected network with $2000$ units in each layer (denoted by $-B$ suffix). 

For CIFAR, the two architecture variants are; $1$) ResNet-18 with $3$ times less feature maps in all the layers (denoted by $-S$ suffix), and 
$2$) Standard ResNet-18 (denoted by $-B$ token).

\begin{figure}[H]
	\begin{subfigure}{0.45\linewidth}
	\begin{center}
		\includegraphics[scale=0.45]{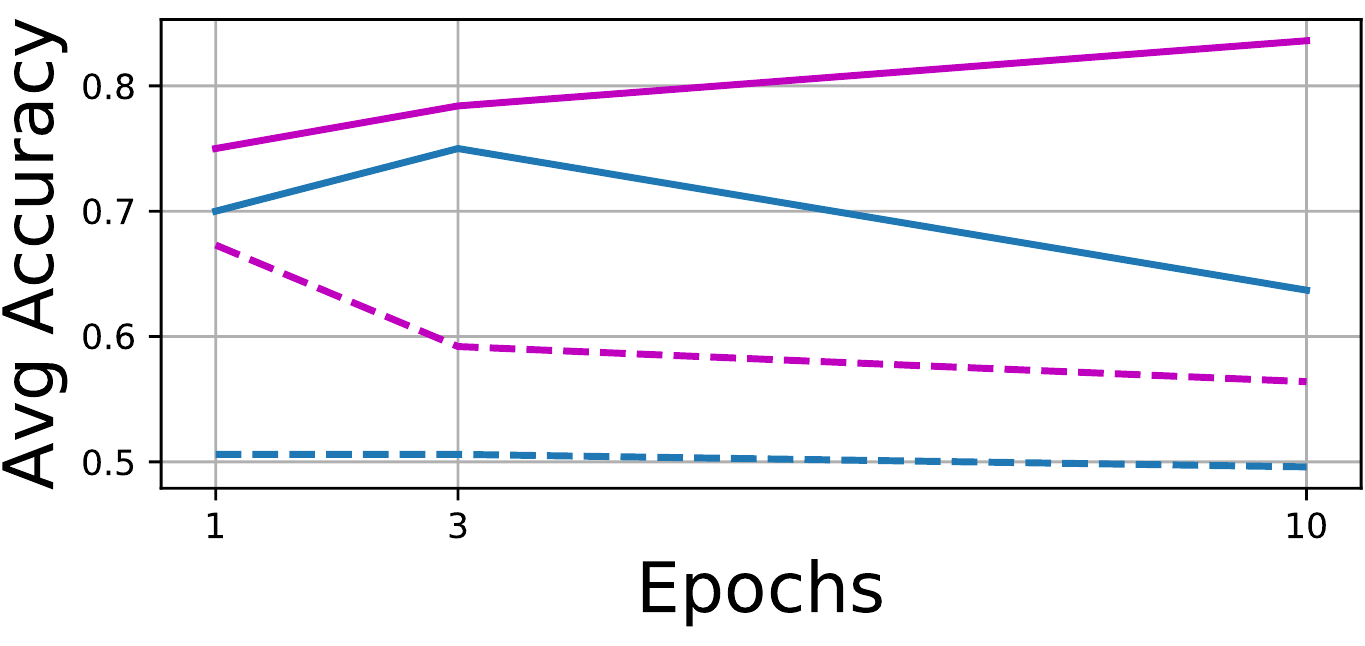}
		\caption{\small Accuracy}
	\end{center}
	\end{subfigure}%
	\begin{subfigure}{0.55\linewidth}
	\begin{center}
		\includegraphics[scale=0.45]{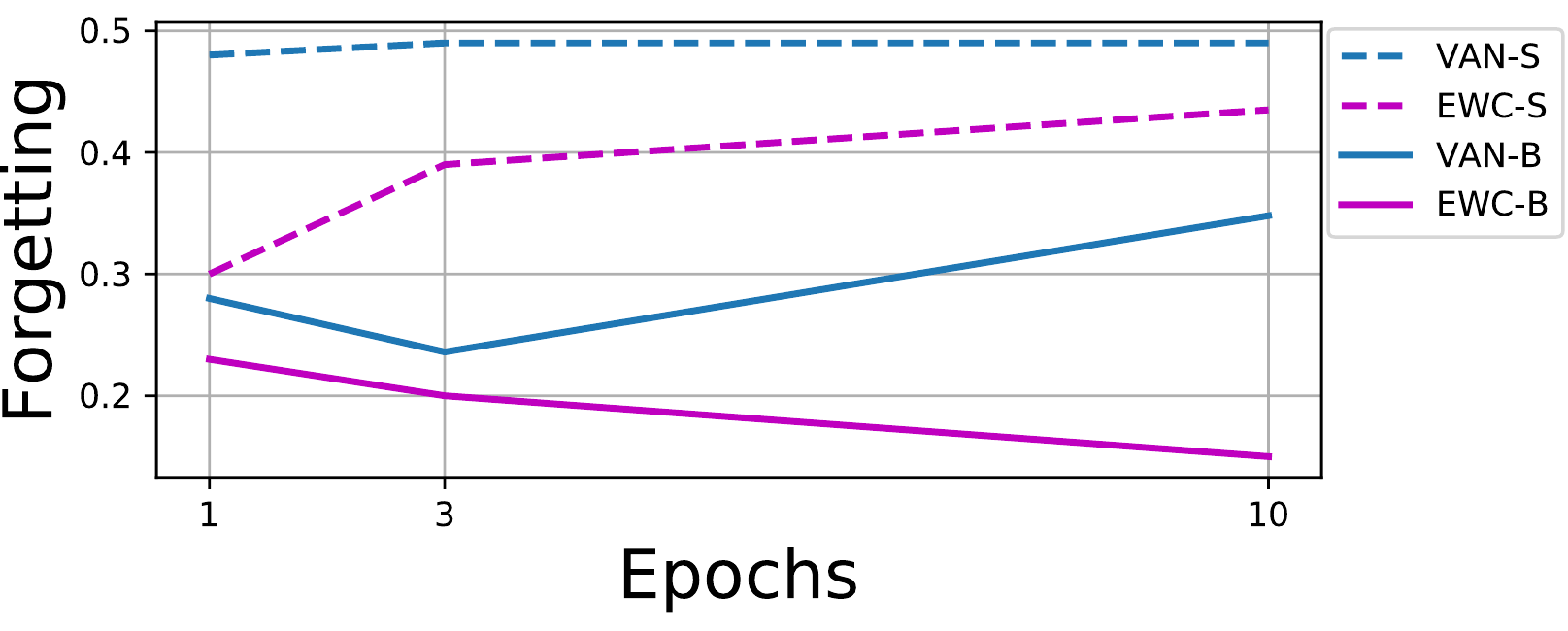}
		\caption{\small Forgetting}
	\end{center}
	\end{subfigure}%
	\vspace{-3mm}
\caption{\em \textbf{Permuted MNIST}: Change in average accuracy and forgetting as the number of epochs are increased. Tokens '-S' and '-B' denote smaller and bigger networks, respectively.}
	\label{fig:supp_mnist_inc_epoch}
\end{figure}

\begin{figure}[H]
	\begin{subfigure}{0.45\linewidth}
	\begin{center}
		\includegraphics[scale=0.45]{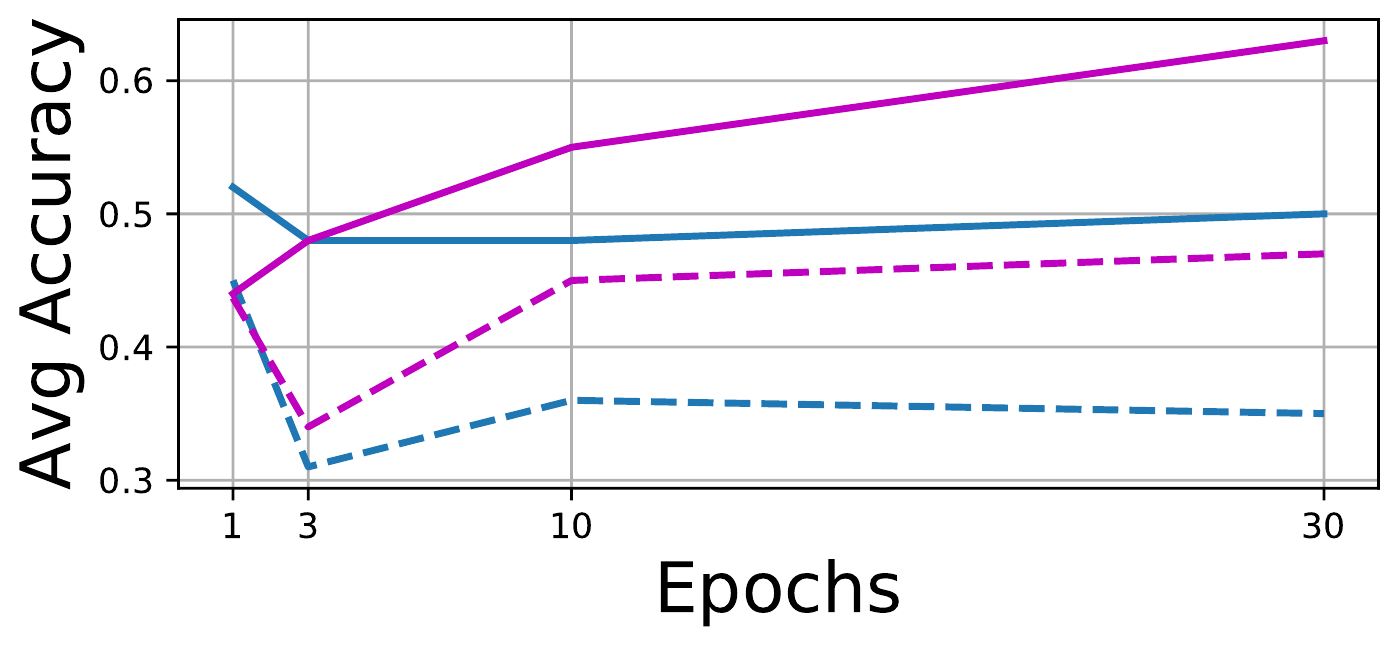}
		\caption{\small Accuracy}
	\end{center}
	\end{subfigure}%
	\begin{subfigure}{0.55\linewidth}
	\begin{center}
		\includegraphics[scale=0.45]{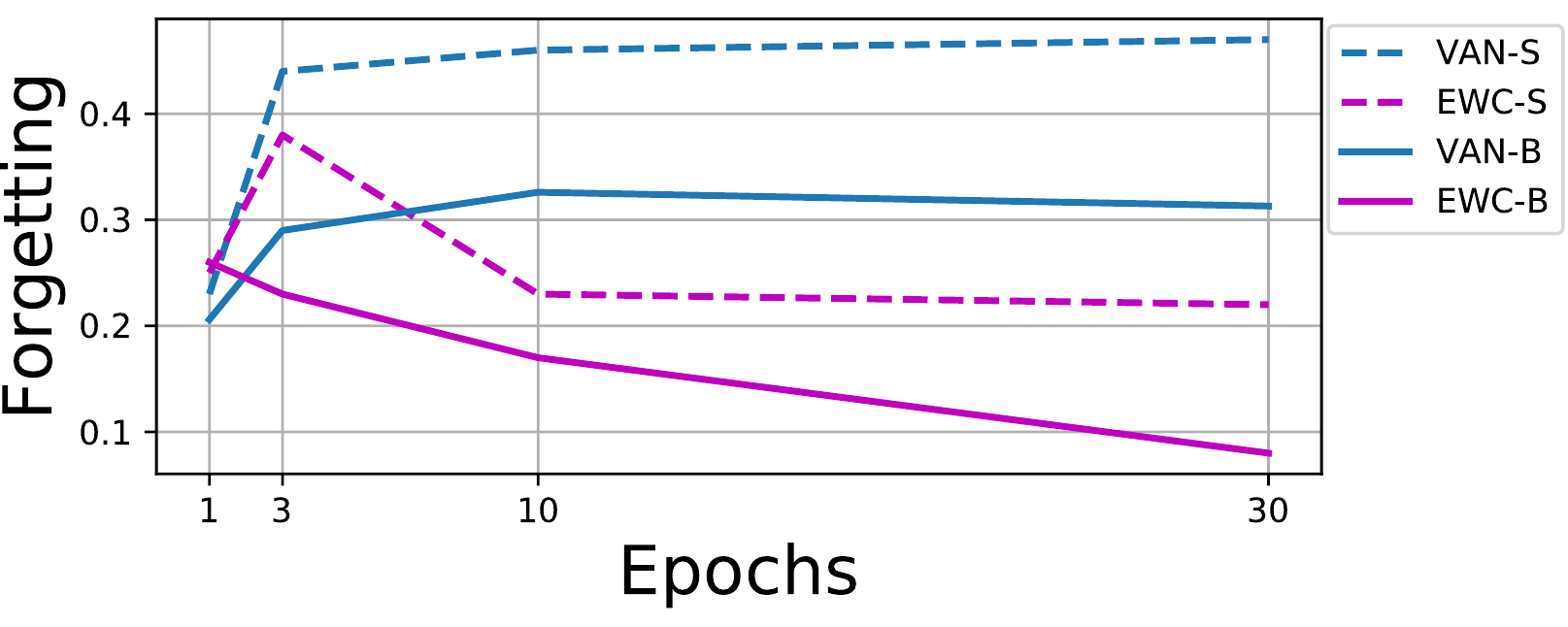}
		\caption{\small Forgetting}
	\end{center}
	\end{subfigure}%
	\vspace{-3mm}
\caption{\em  \textbf{Split CIFAR}: Change in average accuracy and forgetting as the number of epochs are increased. Tokens '-S' and '-B' denote smaller and bigger networks, respectively.}
	\label{fig:supp_cifar_inc_epoch}
\end{figure}

We run the experiments on \van and \ewc\ with increasing the number of epochs from $1$ to $10$ for Permuted MNIST and from $1$ to $30$ for CIFAR. 
For instance, when epoch is set to $10$, it means that the training samples of task $t$ are presented $10$ times before showing examples from task $t+1$.
In Fig.~\ref{fig:supp_mnist_inc_epoch} and~\ref{fig:supp_cifar_inc_epoch} we plot the Average Accuracy (Eq.~\ref{eq:avg_acc}) and Forgetting (Eq.~\ref{eq:fgt}) 
on Permuted MNIST and Split CIFAR, respectively. 

We observe that the average accuracy significantly improves with the number of epochs only when \ewc\ is applied
to the big network. In particular, in the single epoch setting, \ewc peforms similarly to the baseline \van on Split CIFAR which has fewer number of training examples per 
task.

\SKIP{
\begin{figure}[H]
	\centering
	\includegraphics[scale=0.5]{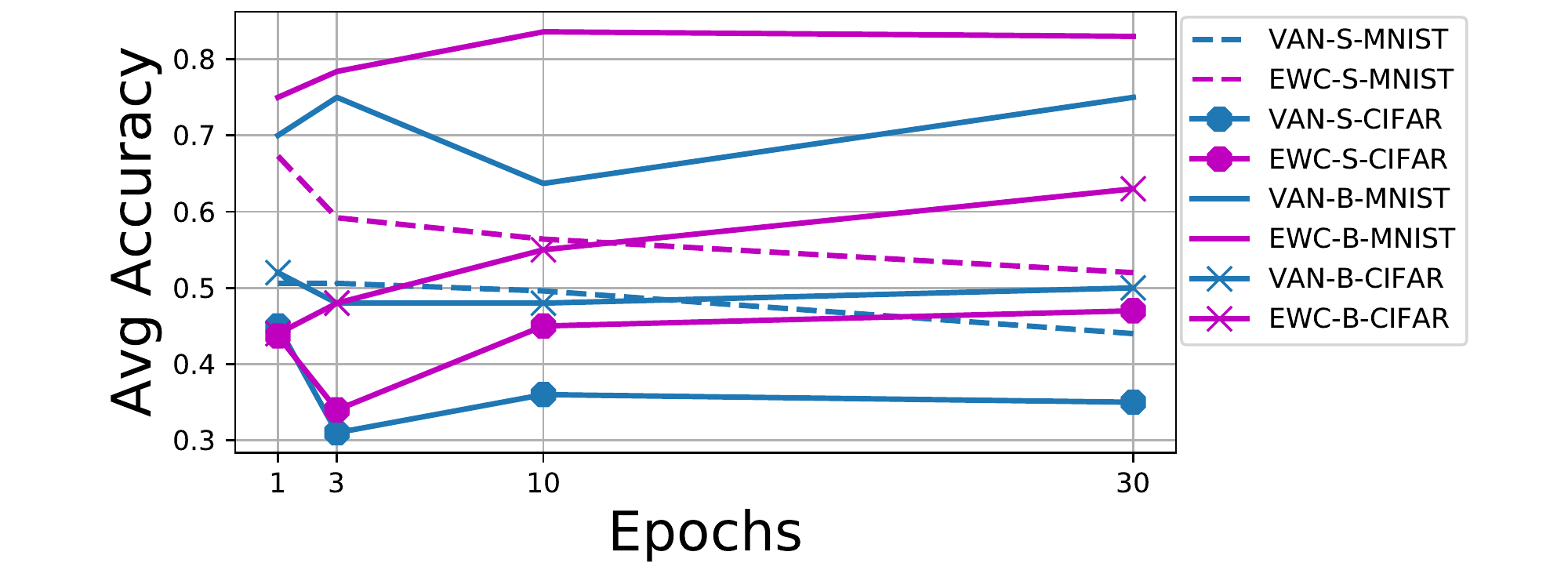}
\end{figure}
}

\section{Hyper-parameter Selection}
\label{sec:supp_hyperparam_selec}
Below we report the hyper-parameters grid considered for different experiments. Note, as described in the Sec.~\ref{sec:experiments} of the main paper, to satisfy the requirement that a learner does not see the data of a task more than once, first $T^{CV}$ tasks are used to cross-validate the hyper-parameters. In all the datasets, the value of $T^{CV}$ is set to `$3$'. The best setting for each experiment is reported in the parenthesis.
\begin{itemize}
        \item {\mtask}
                \begin{itemize}
                        \item learning rate: [$0.3$, $0.1$, $0.03$ (MNIST perm, Split CIFAR, Split CUB, Split AWA), $0.01$, $0.003$, $0.001$, $0.0003$, $0.0001$]
                \end{itemize}
        \item {\mtaskz} 
                \begin{itemize}
                        \item learning rate: [$0.3$, $0.1$, $0.03$ (Split CUB, Split AWA), $0.01$, $0.003$, $0.001$, $0.0003$, $0.0001$]
                \end{itemize}
        \item {\van}
                \begin{itemize}
                        \item learning rate: [$0.3$, $0.1$, $0.03$ (MNIST perm, Split CUB), $0.01$ (Split CIFAR), $0.003$, $0.001$ (Split AWA), $0.0003$, $0.0001$]
                \end{itemize}
        \item {\vanz}
                \begin{itemize}
                                \item learning rate: [$0.3$, $0.1$, $0.03$ (Split CUB), $0.01$, $0.003$ (Split AWA), $0.001$, $0.0003$, $0.0001$]
                \end{itemize}
        \item {\pnn}
                \begin{itemize}
                        \item learning rate: [$0.3$, $0.1$ (MNIST perm, ), $0.03$ (Split CIFAR, Split AWA), $0.01$ (Split CUB), $0.003$, $0.001$, $0.0003$, $0.0001$]
                \end{itemize}
        \item {\ewc}
                \begin{itemize}
                        \item learning rate: [$0.3$, $0.1$, $0.03$ (MNIST perm, Split CIFAR, Split CUB), $0.01$, $0.003$ (Split AWA), $0.001$, $0.0003$, $0.0001$]
                        \item regularization: [$1$ (Split CUB), $10$ (MNIST perm, Split CIFAR), $100$ (Split AWA), $1000$, $10000$]
                \end{itemize}
        \item {\ewcz}
                \begin{itemize}
                        \item learning rate: [$0.3$, $0.1$, $0.03$ (Split CUB), $0.01$, $0.003$ (Split AWA), $0.001$, $0.0003$, $0.0001$]
                        \item regularization: [$1$, $10$ (Split CUB), $100$ (Split AWA), $1000$, $10000$]
                \end{itemize}
        \item {\pii}
                \begin{itemize}
                        \item learning rate: [$0.3$, $0.1$ (MNIST perm), $0.03$ (Split CUB), $0.01$ (Split CIFAR), $0.003$ (Split AWA), $0.001$, $0.0003$, $0.0001$]
                        \item regularization: [$0.001$, $0.01$, $0.1$ (MNIST perm, Split CIFAR, Split CUB), $1$ (Split AWA), $10$]
                \end{itemize}
        \item {\piiz}
                \begin{itemize}
                        \item learning rate: [$0.3$, $0.1$, $0.03$ (Split CUB), $0.01$, $0.003$ (Split AWA), $0.001$, $0.0003$, $0.0001$]
                        \item regularization: [$0.001$, $0.01$, $0.1$ (Split CUB), $1$, $10$ (Split AWA)]
                \end{itemize}
        \item {\mas}
                \begin{itemize}
                        \item learning rate: [$0.3$, $0.1$ (MNIST perm), $0.03$ (Split CIFAR, Split CUB), $0.01$, $0.003$ (Split AWA), $0.001$, $0.0003$, $0.0001$]
                        \item regularization: [$0.01$, $0.1$ (MNIST perm, Split CIFAR, Split CUB), $1$ (Split AWA), $10$]
                \end{itemize}
        \item {\masz}
                \begin{itemize}
                        \item learning rate: [$0.3$, $0.1$, $0.03$ (Split CUB), $0.01$, $0.003$, $0.001$ (Split AWA), $0.0003$, $0.0001$]
                        \item regularization: [$0.01$, $0.1$ (Split CUB, Split AWA), $1$, $10$]
                \end{itemize}
        \item {\rwalk}
                \begin{itemize}
                        \item learning rate: [$0.3$, $0.1$ (MNIST perm), $0.03$ (Split CIFAR, Split CUB), $0.01$, $0.003$ (Split AWA), $0.001$, $0.0003$, $0.0001$]
                        \item regularization: [$0.1$, $1$ (MNIST perm, Split CIFAR, Split CUB), $10$ (Split AWA), $100$, $1000$]
                \end{itemize}
        \item {\rwalkz}
                \begin{itemize}
                        \item learning rate: [$0.3$, $0.1$, $0.03$ (SPLIT CUB), $0.01$, $0.003$ (Split AWA), $0.001$, $0.0003$, $0.0001$]
                                       \item regularization: [$0.1$, $1$ (Split CUB), $10$ (Split AWA), $100$, $1000$]
                \end{itemize}
        \item {\sgem}
                \begin{itemize}
                        \item learning rate: [$0.3$, $0.1$ (MNIST perm), $0.03$ (Split CIFAR, Split CUB), $0.01$ (Split AWA), $0.003$, $0.001$, $0.0003$, $0.0001$]
                \end{itemize}
        \item {\sgemz}
                \begin{itemize}
                        \item learning rate: [$0.3$, $0.1$, $0.03$ (SPLIT CUB), $0.01$, $0.003$ (Split AWA), $0.001$, $0.0003$, $0.0001$]
                \end{itemize}
\end{itemize}

\section{Pictorial Description of Joint Embedding Model }
\begin{figure}[t]
    \centering
    \includegraphics[scale=1.0]{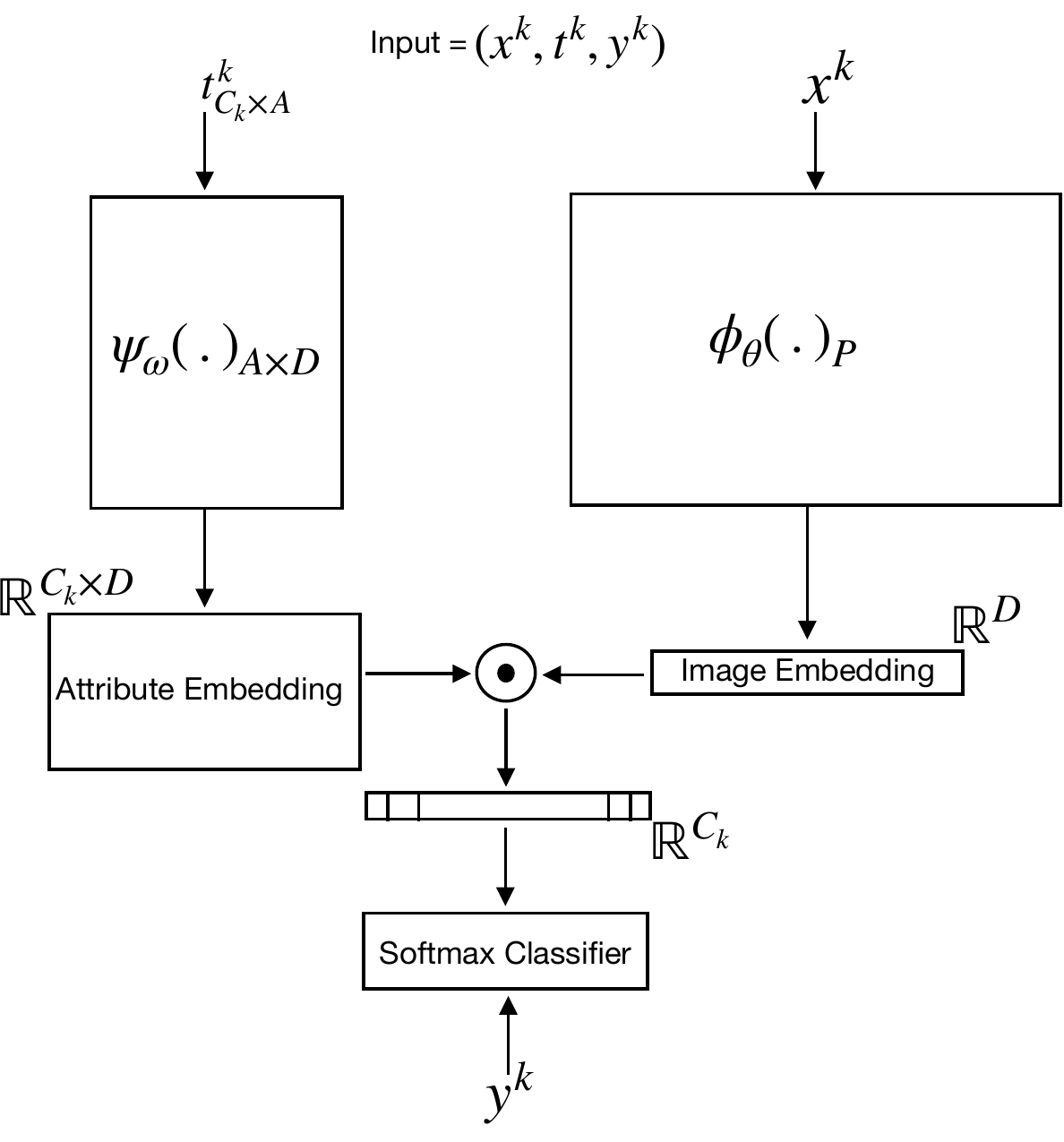}
    \caption{Pictorial description of the joint embedding model discussed in the Sec.~\ref{sec:joint_embed} of the main paper. Modules; $\phi_{\theta}(.)$ and $\psi_{\omega}(.)$ are implemented as feed-forward neural networks with $P$ and $A \times D$ parameters, respectively. The descriptor of task $k$ ($t^k$ ) is a matrix of dimensions $C_k \times A$, shared among all the examples of the task, constructed by concatenating the $A$-dimensional class attribute vectors of $C_k$ classes in the task.}
    \label{fig:je_pict_desc}
\end{figure}

In Fig.~\ref{fig:je_pict_desc} we provide a pictorial description of the joint embedding model discussed in the Sec.~\ref{sec:joint_embed} of the main paper. 
\label{sec:je_pict_desc}
\end{document}